%% file: main.tex
\documentclass{article}

\PassOptionsToPackage{numbers,sort&compress}{natbib}
\usepackage[main,final]{neurips_2026}

\usepackage[utf8]{inputenc}
\usepackage[T1]{fontenc}

\usepackage{amsmath,amssymb,amsfonts}
\usepackage{amsthm}
\usepackage{mathtools}
\usepackage{bm}
\usepackage{nicefrac}

\usepackage{graphicx}
\usepackage{tikz}
\usetikzlibrary{shapes.geometric, matrix, positioning, calc}
\usepackage{pgfplots}
\pgfplotsset{compat=1.18}
\usepgfplotslibrary{groupplots}
\usepackage{subcaption}

\usepackage{booktabs}
\usepackage{multirow}

\usepackage{import}

\usepackage{hyperref}
\usepackage{url}
\usepackage{cleveref}

\usepackage{enumitem}

\usepackage{algorithm}
\usepackage{algorithmic}

\usepackage{microtype}
\usepackage[table]{xcolor}
\usepackage{lipsum}
\newcommand{\WoRD}{\textsc{WoRD}}
\newcommand{\OWR}{\textsc{OWR}}

\newcommand{\unimodal}{unimodal}
\newcommand{\multimodal}{multimodal}

\NewDocumentCommand{\simWP}{m}{\textbf{Sim #1WP}}
\NewDocumentCommand{\simWPc}{s}{%
    \IfBooleanTF{#1}{\textbf{Sim 4WP-c}}{\textbf{Sim 4WP-complex}}}
\NewDocumentCommand{\rounDV}{s}{%
    \IfBooleanTF{#1}{\textbf{rounD-V}}{\textbf{rounD vehicles}}}
\NewDocumentCommand{\indPed}{s}{%
    \IfBooleanTF{#1}{\textbf{inD-Ped}}{\textbf{inD pedestrians}}}

\newcommand{\traj}{\boldsymbol{\tau}}
\newcommand{\obs}{\mathbf{z}}
\newcommand{\state}{\mathbf{x}}

\newcommand{\waypointset}{\mathcal{W}}
\newcommand{\obshistory}{\mathbf{Z}_{1:k}}

\newcommand{\measnoise}{\mathbf{R}}

\newcommand{\noisepredict}{\boldsymbol{\epsilon}_\theta}
\newcommand{\diffstep}{i}
\newcommand{\numsteps}{N}

\newcommand{\R}{\mathbb{R}}

\theoremstyle{plain}

\theoremstyle{definition}

\hypersetup{
    colorlinks=true,
    linkcolor=black,
    citecolor=blue,
    urlcolor=blue,
}

\usepackage{titlesec}
\titlespacing*{\paragraph}{0pt}{0.5ex}{1em}

\title{\WoRD{}: Diffusion-Based Posterior Inference \\ for Online Goal Recognition}

\author{%
  Almog Anschel \\
  CLAIR Lab \\
  Taub Faculty of Computer Science \\
  Technion -- Israel Institute of Technology \\
  Haifa 3200003, Israel \\
  \texttt{anschelalmog@campus.technion.ac.il} \\
  \And
  Sarah Keren \\
  CLAIR Lab \\
  Taub Faculty of Computer Science \\
  Technion -- Israel Institute of Technology \\
  Haifa 3200003, Israel \\
  \texttt{sarahk@technion.ac.il} \\
}

\begin{document}

\maketitle

\input{sections/abstract}

\input{sections/introduction}
\input{sections/related}

\input{sections/problem}
\input{sections/method}

\input{tables/results_main_data}
\input{sections/experiments}
\input{sections/discussion}

\bibliographystyle{plainnat}
\bibliography{references}

\newpage
\appendix
\input{sections/appendix}

\newpage

\end{document}

%% file: sections/abstract.tex
\begin{abstract}
Goal recognition is the task of inferring an agent's intended goal from its observed behavior. Most existing work focuses on high-level symbolic recognition, abstracting away the continuous motion through which agents reach their goal. Methods that do reason over continuous trajectories rely on an explicit dynamics model that is hard to acquire and rarely transfers across settings. Neither approach explicitly learns how agents move toward each candidate goal, limiting accuracy on ambiguous trajectory prefixes where such knowledge is most needed.

We present \WoRD{} (Waypoint Recognition with Diffusion), which uses a learned trajectory prior to infer the agent's goal from partial observations. At each step, \WoRD{} runs one warm-started denoising chain per candidate goal, steers each chain toward the observed trajectory through a closed-form log-likelihood gradient, and reweights the resulting samples by Bayes' rule into a calibrated posterior over goals. We evaluate \WoRD{} on simulated robot settings, pedestrian trajectories, and vehicle trajectories using four metrics targeting distinct failure modes. Across all three settings, the same inference procedure preserves calibrated \multimodal{} beliefs through ambiguous prefixes where filter- and classifier-based observers collapse to a single mode or grow overconfident.
\end{abstract}

%% file: sections/introduction.tex
\section{Introduction}
\label{sec:intro}
Goal recognition is the problem of inferring an agent's intended goal from observations of its behavior. In the online setting, an observer must maintain a belief over candidate goals as observations arrive, before the trajectory becomes unambiguous~\citep{sukthankar_plan_2014,mirsky_introduction_2021}. The problem is hardest early: paths toward adjacent goals overlap until the trajectory commits. An observer that resolves this ambiguity prematurely makes confident-and-wrong predictions, committing to a single hypothesis before evidence warrants it. \Cref{fig:word-overlay-ind} shows a concrete instance: a pedestrian midway through a real intersection where the past trajectory and current heading remain consistent with multiple candidate exits, so a faithful belief must remain \multimodal{}.

Ambiguity of this kind is a structural feature of goal recognition. \citet{keren-goal-2014} characterize this ambiguity via \emph{non-distinctive paths}, prefixes of goal paths before which an observer cannot recognize the agent's goal. A principled observer must maintain \multimodal{} beliefs through this prefix; most existing methods instead collapse the ambiguity before it ends.
Kalman-based filters such as \OWR{}~\citep{guo-online-2024} maintain one Gaussian per candidate goal. When two goals are equally likely, the filter locks onto whichever hypothesis aligns marginally better with early noisy measurements, and the \unimodal{} posterior cannot recover. Particle filters represent richer posteriors, but depend on a well-specified dynamics model. For agents whose motion is goal-directed and obstacle-conditioned, such a model is achievable in principle, but any practical approximation is prone to error and limited in scope~\citep{massardi_error-tolerant_2019}. Discriminative classifiers produce point predictions that do not separate accuracy from calibration. Trained with cross-entropy, they are optimized to place all probability on the labeled goal, pushing probability onto the favored candidate even when observations support multiple goals~\citep{chiari_goal_2023, serina_towards_2025}.

We exploit diffusion models~\citep{ho_denoising_2020, song_score-based_2021}, which learn a data distribution by reversing a gradual noise process: starting from random noise, a learned score function denoises samples step by step into structured outputs. These models are well suited to trajectory data, where the distribution of plausible paths is branching and obstacle-shaped, since a score network can denoise Gaussian noise into coherent, goal-directed paths that no single parametric form captures.

We present \emph{Waypoint Recognition with Diffusion (\WoRD{})}, which uses this denoising framework for inference rather than generation. For each candidate goal in the waypoint set, \WoRD{} runs an independent denoising chain over trajectory space. At each diffusion step, a score-guidance term derived from the observer's measurement model steers the chain toward trajectories consistent with the observations so far. Because the chains share no state, each one keeps its own hypothesis, and the posterior remains \multimodal{} through ambiguous prefixes. Warm-starting from the previous estimate, instead of pure noise, allows the update to track incoming observations in real time. The per-chain trajectory likelihoods are then combined by Bayesian reweighting into a single calibrated posterior over the waypoint set.

We evaluate \WoRD{} on complex simulation environments and on two real-world traffic datasets: rounD~\citep{rounDdataset} (vehicles at roundabouts) and inD~\citep{bock2020ind} (pedestrians at intersections). We compare against five baselines, including a Reverse Diffuser~\citep{janner_planning_2022} baseline that applies the same score network with post-hoc guidance, isolating the contribution of online inference.

\input{figures/word_overview_ind}

\vspace{-0.4cm}
\paragraph{Contributions.}
\begin{enumerate}[leftmargin=*,nosep]
    \item \textbf{Goal recognition as diffusion inference.} We formulate online goal recognition as posterior inference via independent per-goal denoising chains, combined by Bayesian reweighting over reconstructed trajectories, providing a data-driven approach to goal recognition in complex environments where hand-specified motion models are infeasible. Unlike planning-based diffusion methods that condition generation on a known goal~\citep{janner_planning_2022, jiang_motiondiffuser_2023}, \WoRD{} inverts the conditioning: it infers goals from partial observations.
    \item \textbf{Online goal recognition algorithm.} We derive a warm-start mechanism that reuses the previous trajectory estimate as a starting point for each new observation, reducing per-update computation enough for online use at the sensor sampling rate.
    \item \textbf{Analytic observation guidance.} By grounding the
    score-guidance term in the physical geometry of trajectory
    observations, we derive a closed-form update that requires no learned
    classifier and no additional training.
\end{enumerate}

%% file: figures/word_overview_ind.tex
\begin{figure}[!ht]
\centering
\begin{minipage}[c]{0.42\linewidth}
    \caption{%
        \textbf{Online goal recognition at a real intersection
        (illustrative example).}
        A pedestrian (blue dot, with an arrow indicating heading), assumed to be goal-directed, is
        partway along an observed trajectory (solid black) toward one of
        four candidate exits (colored stars); the unobserved future is
     dashed. At this point, the past path and the current
        heading remain consistent with two of the exits (green $w^1$, red $w^2$) so a faithful
        belief over the agent's goal must keep both alive until later
        observations rule one out.
    }%
    \label{fig:word-overlay-ind}
\end{minipage}\hfill
\begin{minipage}[c]{0.55\linewidth}
    \centering
    \includegraphics[width=\linewidth]{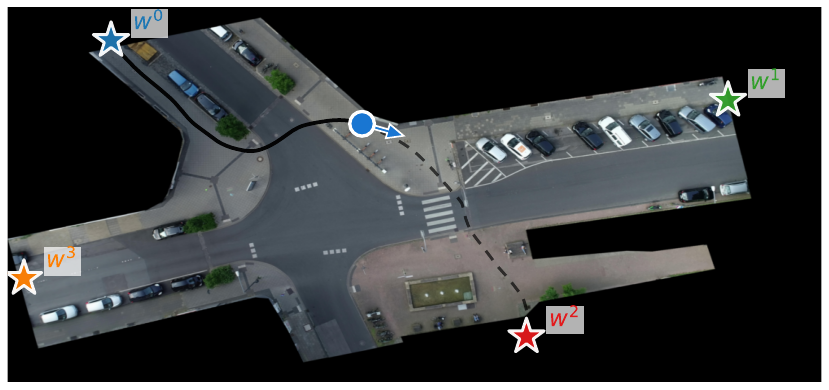}
\end{minipage}
\end{figure}

%% file: sections/related.tex
\section{Related Work}
\label{sec:related}

\paragraph{Goal Recognition:}
Prior taxonomies~\citep{sukthankar_plan_2014, mirsky_introduction_2021, meneguzzi_survey_2021, van-horenbeke-activity-2021} group goal recognition methods into three families, which map to the failure modes of \cref{sec:intro}: planning-based methods relying on hand-crafted domain models, probabilistic filters, and discriminative classifiers.

Planning-based methods cast goal recognition as inverse planning: for each candidate goal, an observer generates an optimal plan and scores it against observed actions~\citep{ramirez_plan_2009, sohrabi_plan_2016}. These methods require a hand-crafted symbolic domain model and assume the agent acts optimally, limiting applicability to continuous or stochastic environments~\citep{kaminka_plan_2018}.
By contrast, \WoRD{} replaces the symbolic domain model with a learned score field trained on observed trajectories, making no assumption about whether the agent plans optimally.

Probabilistic approaches treat goal recognition as online Bayesian inference, with $g$ a candidate goal and $\obshistory$ the trajectory observed so far. They maintain a posterior $p(g \mid \obshistory)$ that updates with each observation~\citep{ramirez_probabilistic_2010}. Particle filters provide a scalable approximation~\citep{massardi_error-tolerant_2019}, but represent beliefs as discrete weighted samples: weight collapse under sharp-peaked likelihoods causes most particles to lose support at the moment the trajectory commits, and downstream gradient-based planning requires lossy smoothing to recover a continuous belief. \WoRD{} avoids these discretizations: the denoising chain represents beliefs as a continuous score field, so observation-likelihood gradients are available analytically at every step without smoothing.

\citet{guo-online-2024} formulate waypoint recognition as a recursive filtering problem, running parallel Kalman filters for each candidate waypoint and weighting them by observation likelihood. This Kalman framework constrains each per-waypoint belief to a \unimodal{} Gaussian, preventing the system from representing ambiguous trajectories where a single partial observation is consistent with multiple divergent paths. \WoRD{} replaces \OWR{}'s per-waypoint filter with a learned score function that maintains the full \multimodal{} posterior.

Learned goal recognition methods cast the problem as discriminative classification~\citep{chiari_goal_2023, serina_towards_2025, shamir_graml_2025, elhadad_general_2026}: a model maps $\obshistory$ to a posterior $p(g \mid \obshistory)$ over candidate goals, bypassing symbolic domain models at the cost of requiring labeled trajectory data. By learning $p(g \mid \obshistory)$ directly without modeling the generative process $p(\obshistory \mid g)$, these methods are brittle when observations are partial or the goal set shifts at test time.
\WoRD{} instead models $p(\obshistory \mid g)$ via a learned trajectory prior, recovering the goal posterior by Bayes' rule and avoiding the train-time overconfidence bias of cross-entropy.

\paragraph{Diffusion Models:}
Score-based generative models learn a data distribution by training a neural network to estimate the score $\nabla_x \log p(x)$ and then draw samples via Langevin dynamics~\citep{song_ermon_2019,song_score-based_2021}. \citet{ho_denoising_2020} show that this framework is equivalent to \emph{denoising diffusion probabilistic models} (DDPMs), unifying the two formulations under a common objective. Conditional generation is typically achieved either through classifier guidance~\citep{dhariwal_diffusion_2021}, which injects the gradient of a separately trained classifier $p(y \mid x_t)$ at sampling time, or through classifier-free guidance~\citep{ho_classifier-free_2022}, which integrates the conditioning into the score network at training time. \WoRD{} belongs to the classifier-guidance family, but replaces the learned classifier with an analytic gradient of the observation likelihood through the measurement model $h$, so no auxiliary classifier is trained or evaluated at inference. The score network itself plays the role of a process model: filter-based observers specify the dynamics analytically, whereas \WoRD{}'s prior is learned from trajectory data.

In planning, \citet{janner_planning_2022} treat trajectory optimization as conditional denoising: a DDPM trained on expert trajectories is guided at inference time toward goal-satisfying samples. MotionDiffuser~\citep{jiang_motiondiffuser_2023} extends this to joint distributions over interacting agent trajectories, recovering \multimodal{} collision-avoidance behaviors that \unimodal{} models cannot represent. In trajectory prediction, conditional diffusion models generate distributions over plausible futures given past motion~\citep{gu_stochastic_2022,shi_singular_2024}, but the latent goal is never explicitly inferred. \WoRD{} inverts this direction: rather than generating trajectories from a known goal or predicting futures from past states, it infers the latent waypoint $w^\star$ from a partial trajectory $\obshistory$ via score-based guidance over a joint trajectory-waypoint distribution.

\paragraph{Motion Forecasting:}
Map-conditioned forecasters such as QCNet~\citep{zhou_query-centric_2023}, HPNet~\citep{tang_hpnet_2024}, and DeMo~\citep{zhang_demo_2024} predict \multimodal{} future trajectories. Since every goal induces a trajectory, forecasting and goal recognition are closely related, yet they report different quantities: a forecaster returns a distribution over future trajectories, whereas a goal recognizer returns a posterior over a given candidate set, which mode probabilities yield only after aggregation over the candidates. The evaluation metrics differ accordingly: FDE, miss rate, and Brier-minFDE~\citep{wilson_argoverse_2021} score the best of $K$ predicted modes, so none of them penalizes a belief that commits to a wrong candidate before the observed prefix disambiguates the goal. Deriving the goal from forecast trajectories is a predict-then-classify strategy; our Reverse Diffuser baseline (\cref{sec:exp-setup}) is a controlled instance of it that shares \WoRD{}'s score network.

%% file: sections/problem.tex
\section{Problem Formulation}
\label{sec:problem}

We consider an agent navigating under closed-loop control in a bounded planar environment $\mathcal{E} \subseteq \R^2$ with static obstacles toward an unknown target $w^* \in \waypointset$, where $\waypointset = \{w^1, \ldots, w^{N}\}$ is a finite set of candidate waypoints known to the observer. At time step $t$, the agent's state is $\state_t = [\mathbf{p}_t^\top, \mathbf{v}_t^\top, \theta_t]^\top \in \mathcal{X} \subseteq \R^5$, with position $\mathbf{p}_t \in \R^2$, velocity $\mathbf{v}_t \in \R^2$, and heading $\theta_t \in [0, 2\pi)$.

The observer cannot access $\state_t$ directly. At each step $k$ it receives a noisy measurement
\begin{equation}
\label{eq:obs-model}
\obs_k = h(\state_k) + \boldsymbol{\eta}_k, \qquad \boldsymbol{\eta}_k \sim \mathcal{N}(\mathbf{0}, \measnoise),
\end{equation}
where $h: \mathcal{X} \to \mathcal{Z} \subseteq \R^m$ is the measurement map and $\measnoise \in \R^{m \times m}$ is a known, time-invariant, symmetric positive-definite noise covariance. The observation history up to time $k$ is $\obshistory = \{\obs_1, \ldots, \obs_k\}$.
The online goal recognition objective is to compute the posterior over candidate goals,
\begin{equation}
\label{eq:objective}
p(w^* = w^j \mid \obshistory), \quad \forall\, w^j \in \waypointset,
\end{equation}
required both to converge to $w^*$ once the agent's behavior is distinctive and to remain \multimodal{} while the prefix is consistent with multiple goals. The observer is online: at each $k$, it must produce \cref{eq:objective} from $\obshistory$ alone, without access to future observations.

We note that optimal plans toward distinct goals may share a long non-distinctive prefix whenever the worst-case distinctiveness (WCD)~\citep{keren-goal-2014} is large. Within such a prefix the true posterior is \multimodal{}, and an observer that collapses to a single mode is wrong in a way the argmax cannot detect, even when the eventual argmax is correct.

\paragraph{Assumptions.}
We make the following assumptions:
(1)~The setting is keyhole: the agent's policy is observer-agnostic, with its actions independent of the observer's presence, and the observer takes no action to influence the agent or environment~\citep{mirsky_introduction_2021};
(2)~the environment is static: obstacles and passages do not change across the observation horizon;
(3)~the observer is trained or configured in the same environment in which it is deployed; no map is provided as input at inference;
(4)~$\waypointset$ is finite and known to the observer;
(5)~$h$ is differentiable; \cref{sec:guidance} uses this to compute likelihood gradients $\nabla_{\state} \log p(\obs_k \mid \state_k)$ analytically;
(6)~the agent is intentional, navigating toward some $w^* \in \waypointset$, though its trajectory may not be optimal due to obstacles or kinematic constraints;
(7)~observations arrive with known timestamps and bounded noise.

\WoRD{} shares assumptions (1)--(4), (6), and (7) with \OWR{} \citep{guo-online-2024}. \OWR{} additionally requires a linear-Gaussian process model for its UKF prediction step, whereas \WoRD{} requires only that $h$ admits a gradient and substitutes a learned score function for the Kalman prediction step.

%% file: sections/method.tex
\section{Waypoint Recognition with Diffusion (\WoRD{})}
\label{sec:method}
\WoRD{} learns the motion through which agents reach their goals from a dataset of demonstrations, removing the need for a hand-specified dynamics model. At inference time we invert the learned prior: at each observation step, one denoising chain per candidate goal is steered toward the observed trajectory, and the chain endpoints are reweighted by Bayes' rule into a posterior.

Given a partial observation history $\obshistory$ and a candidate set $\waypointset = \{w^j\}_{j=1}^N$, \WoRD{} returns a posterior $p(w^j \mid \obshistory)$ over waypoints. \WoRD{} introduces a latent trajectory $\traj = [\state_0, \ldots, \state_{T-1}] \in \R^{T \times d}$ over a fixed horizon $T$. The score network denoises $\traj$ and marginalizes to yield a likelihood per candidate $w^j$; $d$ is the agent state dimension. Throughout, $t$ indexes physical time and $\diffstep$ indexes diffusion step; $\traj_{\diffstep}(t)$ is the value at physical time $t$ of the trajectory at diffusion step $\diffstep$.

For each candidate $w^j$, \WoRD{} models the joint distribution $p(\traj, \obshistory \mid w^j)$ as the product of two terms: a learned \emph{conditional trajectory prior} $p(\traj \mid w^j)$ realized by a DDPM, and an analytic \emph{observation likelihood} $p(\obshistory \mid \traj)$ obtained by differentiating the measurement model $h$ (\cref{eq:obs-model}). At inference, \WoRD{} runs one denoising chain per candidate, guides each reverse step with the analytic observation-likelihood gradient, and reweights the resulting trajectories by Bayes' rule (\cref{eq:mc-likelihood}) to produce the waypoint posterior. \Cref{fig:word-overview-2goals} illustrates the full inference loop.

\subsection{Conditional trajectory prior}
\label{sec:prior}

We model the conditional trajectory distribution $p(\traj \mid w^j)$ for each candidate $w^j \in \waypointset$, treating $w^j \in \R^2$ as a continuous conditioning vector rather than a discrete index, allowing arbitrary waypoint coordinates without architectural changes (architecture in \cref{sec:training}).
The network thus learns the generative direction $p(\traj \mid w)$ and never the discriminative $p(w \mid \traj)$: it has no output indexed by candidate, and the posterior over $\waypointset$ is assembled only at inference by Bayes' rule (\cref{eq:mc-likelihood}), so the candidate set can change at inference without retraining (\cref{app:candidate-sensitivity}).
We adopt a diffusion prior because observation guidance must act online: guidance enters as an additive gradient on the prior score, which requires both a tractable score and an iterative sampler in which to apply it.

\paragraph{Forward process.}
The DDPM~\citep{ho_denoising_2020} forward process corrupts a clean trajectory $\traj_0$ over $\numsteps$ steps using a prescribed variance schedule $\{\beta_{\diffstep}\}_{\diffstep=1}^{\numsteps} \subset (0,1)$:
\begin{equation}
\label{eq:forward}
q(\traj_{\diffstep} \mid \traj_{\diffstep-1}) = \mathcal{N}\!\left(\traj_{\diffstep};\; \sqrt{1-\beta_{\diffstep}}\,\traj_{\diffstep-1},\; \beta_{\diffstep}\mathbf{I}\right),
\end{equation}
with closed-form marginal $q(\traj_{\diffstep} \mid \traj_0) = \mathcal{N}(\traj_{\diffstep};\, \sqrt{\bar{\alpha}_{\diffstep}}\,\traj_0,\, (1{-}\bar{\alpha}_{\diffstep})\mathbf{I})$, where $\bar{\alpha}_{\diffstep} = \prod_{s=1}^{\diffstep}(1{-}\beta_s)$.

\paragraph{Reverse process.}
The reverse process denoises $\traj_{\numsteps} \sim \mathcal{N}(\mathbf{0},\mathbf{I})$ conditioned on goal $w^j$ and observation history $\obshistory$. The conditional mean $\boldsymbol{\mu}_\theta$ is the noise-corrected estimate of $\traj_{\diffstep-1}$: a learned noise predictor $\noisepredict$ estimates the Gaussian perturbation added to $\traj_{\diffstep-1}$ during the forward process at step $\diffstep$, and $\boldsymbol{\mu}_\theta$ subtracts a scaled version of that prediction from $\traj_{\diffstep}$, with coefficients fixed by the noise schedule~\citep{ho_denoising_2020}:
\begin{equation}
\label{eq:reverse}
\begin{aligned}
p_\theta(\traj_{\diffstep-1} \mid \traj_{\diffstep}, w^j, \obshistory) &= \mathcal{N}\!\left(\traj_{\diffstep-1};\; \boldsymbol{\mu}_\theta,\; \sigma_{\diffstep}^2\mathbf{I}\right), \\
\boldsymbol{\mu}_\theta &= \frac{1}{\sqrt{\alpha_{\diffstep}}}\!\left(\traj_{\diffstep} - \frac{\beta_{\diffstep}}{\sqrt{1-\bar{\alpha}_{\diffstep}}}\,\noisepredict\!\left(\traj_{\diffstep}, w^j, \diffstep, \obshistory\right)\right).
\end{aligned}
\end{equation}

\subsection{Training}
\label{sec:training}

We train the score network in two stages that mirror the score decomposition of \cref{eq:score-decomp}: stage~1 fits the trajectory prior $p(\traj \mid w^j)$, and stage~2 fits the observation-conditional $p(\traj \mid w^j, \obshistory)$. Both stages use the same dataset $\mathcal{D} = \{(\traj_0^{(i)}, w^{(i)})\}_{i=1}^{|\mathcal{D}|}$ of trajectories paired with goal waypoints, drawn from the target goal-conditional distribution. End-to-end training tends to be unstable: the network can learn to regress observations directly and bypass the prior, yielding low training loss but poor generalization beyond the observed prefix. Pretraining stage~1 stabilizes optimization and lets each score component be evaluated and ablated independently.

\paragraph{Stage 1: Trajectory prior.}
The noise predictor is trained with the observation channel masked,
\begin{equation}
\label{eq:loss-stage1}
\mathcal{L}_{\text{prior}} = \mathbb{E}_{(\traj_0, w^j) \sim \mathcal{D},\, \boldsymbol{\epsilon},\, \diffstep}\!\left[\bigl\|\boldsymbol{\epsilon} - \noisepredict(\traj_{\diffstep}, w^j, \diffstep, \emptyset)\bigr\|_2^2\right].
\end{equation}

\paragraph{Stage 2: Observation conditioning.}
The pretrained network is fine-tuned end-to-end with the observation channel populated, optimizing the same denoising-score-matching objective with $\obshistory$ in place of $\emptyset$. At each training step we sample a prefix length $k = \lfloor \rho T \rfloor$ with $\rho$ drawn uniformly from a fixed range (in our experiments $\rho \sim \mathcal{U}(0.3,\,0.9)$) and synthesize $\obshistory$ from the clean trajectory $\traj_0$ via the observation model of \cref{eq:obs-model}.

{\bf Architecture.}
The score network is a 2.2M-parameter Temporal U-Net~\citep{janner_planning_2022} with FiLM-based waypoint conditioning~\citep{perez_film_2018}; full architecture, hyperparameters, and the stage~2 data-augmentation schedule are in \cref{app:implementation}.

\subsection{Observation guidance}
\label{sec:guidance}

The observation-conditioned score decomposes via Bayes' rule~\citep{dhariwal_diffusion_2021} (derivation in \cref{app:bayes}):
\begin{equation}
\label{eq:score-decomp}
\nabla_{\traj_{\diffstep}} \log p(\traj_{\diffstep} \mid w^j, \obshistory) = \underbrace{\nabla_{\traj_{\diffstep}} \log p(\traj_{\diffstep} \mid w^j)}_{\text{learned prior}} + \underbrace{\nabla_{\traj_{\diffstep}} \log p(\obshistory \mid \traj_{\diffstep})}_{\text{observation guidance}}.
\end{equation}
In practice we evaluate the leading term with the stage-2 conditional network (\cref{sec:training}) rather than the bare prior, as $-\noisepredict(\traj_{\diffstep}, w^j, \diffstep, \obshistory) / \sqrt{1-\bar{\alpha}_{\diffstep}}$ via the noise-prediction-to-score equivalence~\citep{ho_denoising_2020}. We replace the Bayes-rule weight on the guidance term by a tunable scale $\lambda$~\citep{dhariwal_diffusion_2021}, which increases observation fidelity beyond what the stage-2 conditioning already captures; sensitivity is reported in \cref{app:ablations}. For the guidance term, the Gaussian observation model from \cref{eq:obs-model} yields a closed-form log-likelihood, differentiable in $\traj_{\diffstep}$ through $h$:
\begin{equation}
\label{eq:obs-likelihood}
\log p(\obshistory \mid \traj_{\diffstep}) = -\frac{1}{2}\sum_{t=1}^{k}\left\|\obs_t - h\!\left(\traj_{\diffstep}(t)\right)\right\|_{\measnoise^{-1}}^{2}.
\end{equation}
The guidance gradient $\mathbf{g}^j = \nabla_{\traj_{\diffstep}^j} \log p(\obshistory \mid \traj_{\diffstep}^j)$ is computed by backpropagation through $h$, which assumption~5 (\cref{sec:problem}) makes differentiable, so no learned likelihood model or classifier head~\citep{dhariwal_diffusion_2021} is needed. Together with the network's learned observation conditioning (\cref{sec:training}), the analytic gradient provides an exact likelihood signal at every diffusion step.

\subsection{Online inference with warm-start}
\label{sec:warmstart}

Denoising from pure noise at every observation is incompatible with online operation. The previous estimate $\hat{\traj}_{k-1}$ (cached at step~4 below) already lies near the data manifold, so we re-noise it to an intermediate step $\diffstep^* \ll \numsteps$ via the closed-form marginal,
\begin{equation}
\label{eq:warmstart}
\traj_{\diffstep^*}^j \sim q(\traj_{\diffstep^*} \mid \hat{\traj}_{k-1}) = \mathcal{N}\!\left(\sqrt{\bar{\alpha}_{\diffstep^*}}\,\hat{\traj}_{k-1},\,(1{-}\bar{\alpha}_{\diffstep^*})\mathbf{I}\right),
\end{equation}
and reverse from $\diffstep^*$ to $0$, cutting per-update cost by a factor of $\numsteps / \diffstep^*$. At each new observation $\obs_k$, \WoRD{} performs four steps per candidate $w^j$:
\begin{enumerate}[leftmargin=2em,itemsep=2pt,topsep=2pt]
\item \textbf{Re-noise}: corrupt $\hat{\traj}_{k-1}^j$ to step $\diffstep^*$.
\item \textbf{Guided reverse}: denoise from $\diffstep^*$ to $0$, adding $\lambda \mathbf{g}^j$ to the prior score at each step.
\item \textbf{Reweight}: update the posterior via \cref{eq:mc-likelihood}.
\item \textbf{Store}: cache $\hat{\traj}_k^j = \traj_0^j$ for the next warm-start.
\end{enumerate}
To reduce variance, we draw $S$ trajectories per candidate from the shared warm-start and estimate the marginal likelihood as
\begin{equation}
\label{eq:mc-likelihood}
\hat{p}(\obshistory \mid w^j) \;=\; \frac{1}{S} \sum_{s=1}^{S} p\!\left(\obshistory \mid \traj_0^{j,(s)}\right).
\end{equation}
Combining these per-candidate marginal likelihoods with the candidate prior via Bayes' rule yields the posterior $p(w^j \mid \obshistory)$. \Cref{app:algorithm} states the full inference procedure.

\input{figures/word_overview_2goals}

%% file: figures/word_overview_2goals.tex
\providecommand{\cref}[1]{\ref{#1}}

\def\BoardW{10}   \def\BoardH{10}
\def\WPoneX{2}    \def\WPoneY{8.5}
\def\WPtwoX{8}    \def\WPtwoY{8.5}
\def\ObsX{5}      \def\ObsY{5}
\def\ObsR{1.5}
\def\SX{5}        \def\SY{1}

\tikzset{
    gridscale/.style={x=0.38cm, y=0.38cm},
    noisetraj/.style={gray!40, line width=0.4pt, opacity=0.55},
    guidearrow/.style={->, orange!80!red, line width=1.4pt, line cap=round},
}

\colorlet{cTrue}{blue!80!black}        \colorlet{cTrueLight}{blue!40}
\colorlet{cDist}{red!70!black}         \colorlet{cDistLight}{red!40}

\newcommand{\drawboard}{%
    \draw[gray!15, very thin] (0,0) grid[step=2] (\BoardW,\BoardH);
    \fill[gray!30] (\ObsX,\ObsY) circle (\ObsR);
    \draw[black!60, thick] (\ObsX,\ObsY) circle (\ObsR);
    \node[cTrue, star, star points=5, star point ratio=2.2,
          fill=blue!20, draw=cTrue, line width=0.5pt,
          minimum size=8pt, inner sep=0pt] at (\WPoneX,\WPoneY) {};
    \node[cTrue, font=\tiny\bfseries, anchor=south]
          at (\WPoneX,\WPoneY+0.25) {$w^1$};
    \fill[cDist] (\WPtwoX,\WPtwoY) circle (0.12);
    \draw[cDist, line width=0.5pt] (\WPtwoX,\WPtwoY) circle (0.12);
    \node[cDist, font=\tiny\bfseries, anchor=south]
          at (\WPtwoX,\WPtwoY+0.25) {$w^2$};
    \draw[black, thick, fill=white] (\SX,\SY) circle (0.1);
    \node[font=\tiny, anchor=north] at (\SX,0.7) {Start};
}

\newcommand{\probbox}[2]{%
    \par\vspace{2pt}%
    \parbox{0.95\linewidth}{\scriptsize\centering
        $P(w^1|\mathbf{Z})=#1$\quad$P(w^2|\mathbf{Z})=#2$}%
}

\newcommand{\obsmark}[2]{%
    \draw[black!70, line width=0.6pt]
        (#1-0.18,#2-0.18) -- (#1+0.18,#2+0.18)
        (#1-0.18,#2+0.18) -- (#1+0.18,#2-0.18);
}

\newcommand{\drawObsShort}{%
    \obsmark{5.10}{1.40}
    \obsmark{4.70}{1.92}
    \obsmark{4.85}{2.42}
    \obsmark{4.30}{3.10}
    \obsmark{4.15}{3.42}
}

\newcommand{\drawObsMed}{%
    \obsmark{5.10}{1.40}
    \obsmark{4.70}{1.92}
    \obsmark{4.85}{2.42}
    \obsmark{4.30}{3.10}
    \obsmark{4.15}{3.42}
    \obsmark{3.40}{3.92}
    \obsmark{3.40}{4.42}
    \obsmark{2.95}{4.92}
}

\newcommand{\agentdot}[3]{%
    \fill[green!60!black] (#1,#2) circle (0.18);
    \draw[->, green!60!black, line width=1.3pt]
        (#1,#2) -- +({0.55*cos(#3)},{0.55*sin(#3)});
}

\newcommand{\pathshort}{%
    \draw[black, line width=1.2pt]
        (5,1) -- (4.95,1.5) -- (4.85,2.0) -- (4.70,2.5)
        -- (4.45,3.0) -- (4.00,3.50);
}
\newcommand{\pathmed}{%
    \draw[black, line width=1.2pt]
        (5,1) -- (4.95,1.5) -- (4.85,2.0) -- (4.70,2.5)
        -- (4.45,3.0) -- (4.00,3.50) -- (3.55,4.00)
        -- (3.25,4.50) -- (3.10,5.00);
}
\newcommand{\pathfull}{%
    \draw[black, line width=1.2pt]
        (5,1) -- (4.95,1.5) -- (4.85,2.0) -- (4.70,2.5)
        -- (4.45,3.0) -- (4.00,3.50) -- (3.55,4.00)
        -- (3.25,4.50) -- (3.10,5.00) -- (3.00,5.50)
        -- (2.85,6.00) -- (2.70,6.50) -- (2.55,7.00)
        -- (2.40,7.50) -- (2.25,8.00);
}

\newcommand{\drawPriorClusters}{%
    \draw[cTrueLight, line width=0.5pt, opacity=0.55]
        (5,1) .. controls (4.0,3.0) and (2.5,5.5) .. (2.10,8.40);
    \draw[cTrueLight, line width=0.5pt, opacity=0.45]
        (5,1) .. controls (3.5,2.5) and (2.0,5.0) .. (1.85,8.55);
    \draw[cTrueLight, line width=0.5pt, opacity=0.50]
        (5,1) .. controls (4.0,3.5) and (2.5,6.5) .. (2.30,8.30);
    \draw[cDistLight, line width=0.5pt, opacity=0.55]
        (5,1) .. controls (6.0,3.0) and (7.5,5.5) .. (7.90,8.40);
    \draw[cDistLight, line width=0.5pt, opacity=0.45]
        (5,1) .. controls (6.5,2.5) and (8.0,5.0) .. (8.15,8.55);
    \draw[cDistLight, line width=0.5pt, opacity=0.50]
        (5,1) .. controls (6.0,3.5) and (7.5,6.5) .. (7.70,8.30);
    \draw[gray!25, line width=0.3pt, opacity=0.3]
        (5,1) .. controls (5.0,2.5) and (3.0,3.5) .. (2.7,5.0);
    \draw[gray!25, line width=0.3pt, opacity=0.3]
        (5,1) .. controls (5.0,2.5) and (7.0,3.5) .. (7.3,5.0);
}

\newcommand{\drawAmbigClusters}{%
    \draw[cTrueLight, line width=0.55pt, opacity=0.60]
        (4.00,3.50) .. controls (3.0,5.0) and (2.5,7.0) .. (2.10,8.40);
    \draw[cTrueLight, line width=0.55pt, opacity=0.50]
        (4.00,3.50) .. controls (2.7,5.0) and (2.0,7.0) .. (1.85,8.55);
    \draw[cTrueLight, line width=0.55pt, opacity=0.55]
        (4.00,3.50) .. controls (2.8,5.5) and (2.5,7.5) .. (2.30,8.30);
    \draw[cDistLight, line width=0.5pt, opacity=0.50]
        (4.00,3.50) .. controls (1.0,7.5) and (7.5,9.5) .. (7.90,8.40);
    \draw[cDistLight, line width=0.5pt, opacity=0.42]
        (4.00,3.50) .. controls (0.5,8.0) and (7.0,9.5) .. (8.15,8.55);
    \draw[cDistLight, line width=0.5pt, opacity=0.45]
        (4.00,3.50) .. controls (1.5,7.0) and (8.0,9.0) .. (7.70,8.30);
}

\newcommand{\drawGuidedSamples}{%
    \draw[cTrueLight, line width=0.7pt, opacity=0.80]
        (3.10,5.00) .. controls (2.7,6.3) and (2.3,7.4) .. (2.10,8.40);
    \draw[cTrueLight, line width=0.7pt, opacity=0.70]
        (3.10,5.00) .. controls (2.5,6.2) and (2.0,7.5) .. (1.85,8.55);
    \draw[cTrueLight, line width=0.7pt, opacity=0.75]
        (3.10,5.00) .. controls (2.9,6.4) and (2.5,7.5) .. (2.30,8.30);
    \draw[cDistLight, line width=0.5pt, opacity=0.40]
        (3.10,5.00) .. controls (3.0,8.0) and (6.5,8.8) .. (7.90,8.40);
    \draw[cDistLight, line width=0.5pt, opacity=0.30]
        (3.10,5.00) .. controls (2.5,8.5) and (6.5,9.0) .. (8.10,8.50);
}

\newcommand{\drawGuidanceArrows}{%
    \draw[guidearrow] (1.80,5.00) -- (2.72,5.00);
    \draw[guidearrow] (3.10,3.90) -- (3.10,4.62);
    \draw[guidearrow] (3.70,5.80) -- (3.30,5.20);
    \node[font=\tiny, orange!80!red, anchor=east] at (2.40,4.40)
        {$\nabla_{\traj}\!\log p(\mathbf{Z}|\traj)$};
}

\newcommand{\drawPosterior}{%
    \draw[cTrue, line width=1.4pt, opacity=0.85]
        (2.25,8.00) .. controls (2.20,8.20) and (2.10,8.40) .. (2.05,8.45);
    \draw[cDistLight, line width=0.6pt, opacity=0.18]
        (2.25,8.00) .. controls (4.0,8.5) and (6.5,8.6) .. (7.95,8.45);
}

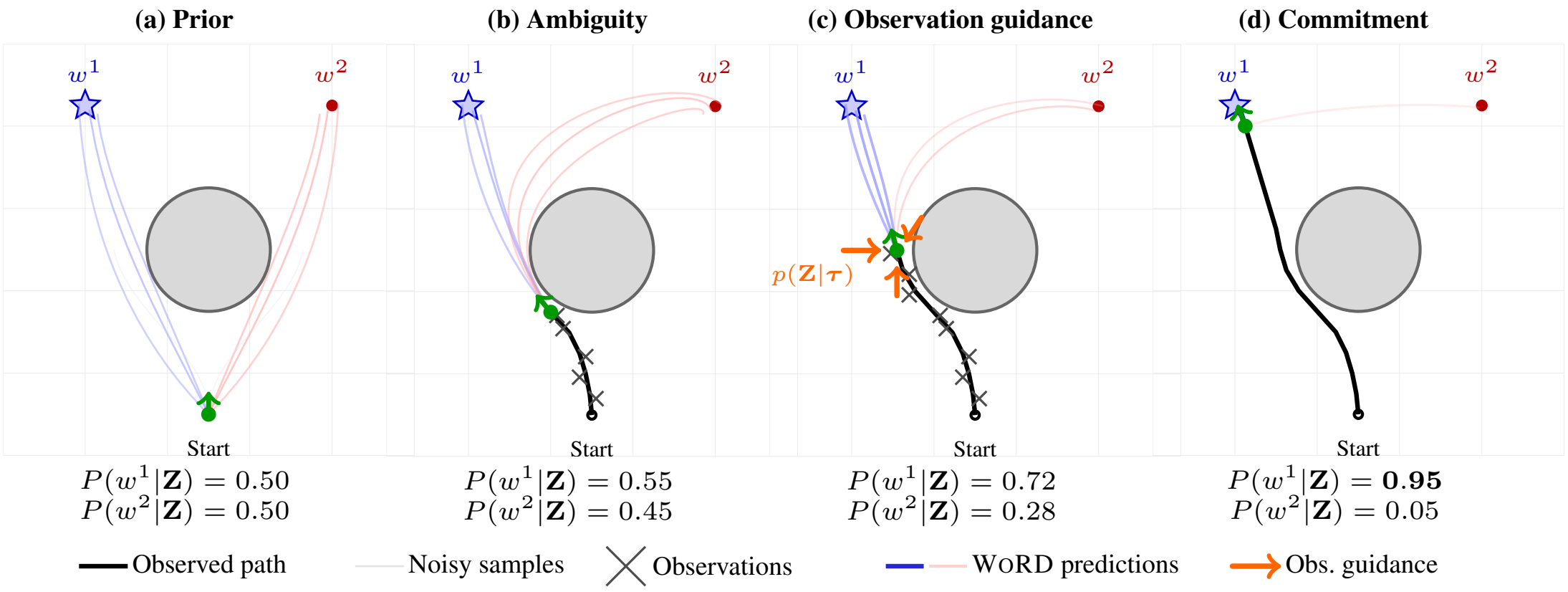
\begin{figure*}[th]
\centering

\begin{subfigure}[t]{0.24\textwidth}
\centering
\textbf{\scriptsize (a) Prior}\\[2pt]
\begin{tikzpicture}[gridscale]
\begin{scope}\clip (0,0) rectangle (\BoardW,\BoardH);
\drawboard
\drawPriorClusters
\agentdot{\SX}{\SY}{90}
\end{scope}
\end{tikzpicture}
\probbox{0.50}{0.50}
\end{subfigure}\hfill
\begin{subfigure}[t]{0.24\textwidth}
\centering
\textbf{\scriptsize (b) Ambiguity}\\[2pt]
\begin{tikzpicture}[gridscale]
\begin{scope}\clip (0,0) rectangle (\BoardW,\BoardH);
\drawboard
\drawAmbigClusters
\pathshort
\drawObsShort
\agentdot{4.00}{3.50}{130}
\end{scope}
\end{tikzpicture}
\probbox{0.55}{0.45}
\end{subfigure}\hfill
\begin{subfigure}[t]{0.24\textwidth}
\centering
\textbf{\scriptsize (c) Observation guidance}\\[2pt]
\begin{tikzpicture}[gridscale]
\begin{scope}\clip (0,0) rectangle (\BoardW,\BoardH);
\drawboard
\drawGuidedSamples
\pathmed
\drawObsMed
\agentdot{3.10}{5.00}{105}
\drawGuidanceArrows
\end{scope}
\end{tikzpicture}
\probbox{0.72}{0.28}
\end{subfigure}\hfill
\begin{subfigure}[t]{0.24\textwidth}
\centering
\textbf{\scriptsize (d) Commitment}\\[2pt]
\begin{tikzpicture}[gridscale]
\begin{scope}\clip (0,0) rectangle (\BoardW,\BoardH);
\drawboard
\drawPosterior
\pathfull
\agentdot{2.25}{8.00}{110}
\end{scope}
\end{tikzpicture}
\probbox{\mathbf{0.95}}{0.05}
\end{subfigure}

\vspace{4pt}
\noindent\makebox[\textwidth][c]{%
\begin{tikzpicture}[every node/.style={font=\scriptsize, inner sep=1pt}]
    \draw[black, line width=1.2pt] (0,0) -- (0.45,0);
    \node[anchor=west] at (0.45,0) {Observed path};
    \draw[gray!40, line width=0.5pt, opacity=0.55] (2.55,0) -- (3.0,0);
    \node[anchor=west] at (3.0,0) {Noisy samples};
    \obsmark{5.05}{0}
    \node[anchor=west, inner sep=2pt] at (5.22,0) {Observations};
    \draw[cTrue,      line width=1.2pt, opacity=0.85] (7.45,0) -- (7.80,0);
    \draw[cDistLight, line width=0.8pt, opacity=0.45] (7.85,0) -- (8.20,0);
    \node[anchor=west] at (8.20,0) {\WoRD{} predictions};
    \draw[guidearrow] (10.65,0) -- (11.10,0);
    \node[anchor=west] at (11.10,0) {Obs.\ guidance};
\end{tikzpicture}%
}

\caption{%
    \textbf{\WoRD{} online inference on a 2-goal synthetic scenario.}
    An agent (green) starts at the bottom of a $10\times10$ world and
    must reach one of two goals separated by a circular obstacle;
    \WoRD{} maintains one denoising chain per candidate goal and
    updates a posterior as observations arrive.
    \textbf{(a)}~\emph{Prior.} Before any observation, the score network
    $\noisepredict$ shapes Gaussian noise into two obstacle-aware
    trajectory clusters, one towards each waypoint, and the goal prior is uniform.
    \textbf{(b)}~\emph{Ambiguity.} The agent has skirted the lower
    edge of the obstacle but its heading is still consistent with both
    goals; both clusters survive and the posterior remains nearly
    balanced. A filtering baseline would already have
    collapsed onto one mode here.
    \textbf{(c)}~\emph{Observation guidance.} Once the agent commits
    left of the obstacle, the score-guidance term
    $\nabla_{\traj} \log p(\mathbf{Z} \mid \traj)$ (orange) pins each
    chain to the observed path; the right cluster must arc awkwardly
    over the obstacle to remain reachable, and posterior mass shifts
    toward $w^1$.
    \textbf{(d)}~\emph{Commitment.} As the agent approaches $w^1$, the
    surviving chain dominates and the posterior collapses. Pseudocode is in \cref{alg:word}.}
\label{fig:word-overview-2goals}
\end{figure*}

%% file: tables/results_main_data.tex
\newcommand{\dNaiveTopOneSimThree}{$64.2{\pm}2.1$}
\newcommand{\dNaiveTopOneSimFour}{$55.6{\pm}2.4$}
\newcommand{\dNaiveTopOneSimCplx}{$38.4{\pm}3.2$}
\newcommand{\dNaiveTopOneRnD}{$32.8{\pm}5.5$}
\newcommand{\dNaiveTopOneIPed}{$27.4{\pm}6.2$}

\newcommand{\dOWRTopOneSimThree}{$81.4{\pm}1.6$}
\newcommand{\dOWRTopOneSimFour}{$75.2{\pm}1.9$}
\newcommand{\dOWRTopOneSimCplx}{$58.7{\pm}2.7$}
\newcommand{\dOWRTopOneRnD}{$53.1{\pm}4.7$}
\newcommand{\dOWRTopOneIPed}{$31.4{\pm}6.2$}

\newcommand{\dPFTopOneSimThree}{$74.8{\pm}2.1$}
\newcommand{\dPFTopOneSimFour}{$76.8{\pm}2.0$}
\newcommand{\dPFTopOneSimCplx}{$64.2{\pm}2.4$}
\newcommand{\dPFTopOneRnD}{$62.7{\pm}4.5$}
\newcommand{\dPFTopOneIPed}{$54.2{\pm}5.2$}

\newcommand{\dGRUTopOneSimThree}{{\boldmath $88.4{\pm}1.3$}}
\newcommand{\dGRUTopOneSimFour}{{\boldmath $82.9{\pm}1.4$}}
\newcommand{\dGRUTopOneSimCplx}{$67.8{\pm}2.1$}
\newcommand{\dGRUTopOneRnD}{$63.4{\pm}4.6$}
\newcommand{\dGRUTopOneIPed}{$57.1{\pm}5.4$}

\newcommand{\dRDTopOneSimThree}{$84.3{\pm}1.4$}
\newcommand{\dRDTopOneSimFour}{$80.7{\pm}1.6$}
\newcommand{\dRDTopOneSimCplx}{$70.4{\pm}1.9$}
\newcommand{\dRDTopOneRnD}{$64.9{\pm}4.4$}
\newcommand{\dRDTopOneIPed}{$58.6{\pm}5.1$}

\newcommand{\dWoRDTopOneSimThree}{$84.6{\pm}1.5$}
\newcommand{\dWoRDTopOneSimFour}{$82.4{\pm}1.3$}
\newcommand{\dWoRDTopOneSimCplx}{{\boldmath $78.2{\pm}1.6$}}
\newcommand{\dWoRDTopOneRnD}{{\boldmath $71.3{\pm}3.7$}}
\newcommand{\dWoRDTopOneIPed}{{\boldmath $65.7{\pm}5.1$}}

\newcommand{\dNaiveBrierSimThree}{$0.41{\pm}0.03$}
\newcommand{\dNaiveBrierSimFour}{$0.49{\pm}0.03$}
\newcommand{\dNaiveBrierSimCplx}{$0.59{\pm}0.04$}
\newcommand{\dNaiveBrierRnD}{$0.63{\pm}0.06$}
\newcommand{\dNaiveBrierIPed}{$0.67{\pm}0.08$}

\newcommand{\dOWRBrierSimThree}{$0.17{\pm}0.02$}
\newcommand{\dOWRBrierSimFour}{$0.24{\pm}0.03$}
\newcommand{\dOWRBrierSimCplx}{$0.34{\pm}0.03$}
\newcommand{\dOWRBrierRnD}{$0.42{\pm}0.06$}
\newcommand{\dOWRBrierIPed}{$0.48{\pm}0.07$}

\newcommand{\dPFBrierSimThree}{$0.19{\pm}0.02$}
\newcommand{\dPFBrierSimFour}{$0.23{\pm}0.02$}
\newcommand{\dPFBrierSimCplx}{$0.30{\pm}0.03$}
\newcommand{\dPFBrierRnD}{$0.31{\pm}0.05$}
\newcommand{\dPFBrierIPed}{$0.36{\pm}0.05$}

\newcommand{\dGRUBrierSimThree}{{\boldmath $0.12{\pm}0.01$}}
\newcommand{\dGRUBrierSimFour}{{\boldmath $0.16{\pm}0.02$}}
\newcommand{\dGRUBrierSimCplx}{$0.27{\pm}0.02$}
\newcommand{\dGRUBrierRnD}{$0.32{\pm}0.05$}
\newcommand{\dGRUBrierIPed}{$0.39{\pm}0.06$}

\newcommand{\dRDBrierSimThree}{$0.14{\pm}0.01$}
\newcommand{\dRDBrierSimFour}{$0.18{\pm}0.02$}
\newcommand{\dRDBrierSimCplx}{$0.23{\pm}0.02$}
\newcommand{\dRDBrierRnD}{$0.27{\pm}0.05$}
\newcommand{\dRDBrierIPed}{$0.31{\pm}0.05$}

\newcommand{\dWoRDBrierSimThree}{$0.13{\pm}0.01$}
\newcommand{\dWoRDBrierSimFour}{$0.17{\pm}0.02$}
\newcommand{\dWoRDBrierSimCplx}{{\boldmath $0.19{\pm}0.02$}}
\newcommand{\dWoRDBrierRnD}{{\boldmath $0.22{\pm}0.04$}}
\newcommand{\dWoRDBrierIPed}{{\boldmath $0.25{\pm}0.04$}}

\newcommand{\dNaiveCVaRSimThree}{$0.06{\pm}0.01$}
\newcommand{\dNaiveCVaRSimFour}{$0.04{\pm}0.01$}
\newcommand{\dNaiveCVaRSimCplx}{$0.03{\pm}0.01$}
\newcommand{\dNaiveCVaRRnD}{$0.02{\pm}0.02$}
\newcommand{\dNaiveCVaRIPed}{$0.02{\pm}0.02$}

\newcommand{\dOWRCVaRSimThree}{$0.18{\pm}0.02$}
\newcommand{\dOWRCVaRSimFour}{$0.13{\pm}0.02$}
\newcommand{\dOWRCVaRSimCplx}{$0.08{\pm}0.01$}
\newcommand{\dOWRCVaRRnD}{$0.06{\pm}0.02$}
\newcommand{\dOWRCVaRIPed}{$0.05{\pm}0.02$}

\newcommand{\dPFCVaRSimThree}{$0.26{\pm}0.02$}
\newcommand{\dPFCVaRSimFour}{$0.22{\pm}0.02$}
\newcommand{\dPFCVaRSimCplx}{$0.17{\pm}0.02$}
\newcommand{\dPFCVaRRnD}{$0.14{\pm}0.03$}
\newcommand{\dPFCVaRIPed}{$0.11{\pm}0.03$}

\newcommand{\dGRUCVaRSimThree}{$0.13{\pm}0.02$}
\newcommand{\dGRUCVaRSimFour}{$0.10{\pm}0.02$}
\newcommand{\dGRUCVaRSimCplx}{$0.07{\pm}0.01$}
\newcommand{\dGRUCVaRRnD}{$0.05{\pm}0.02$}
\newcommand{\dGRUCVaRIPed}{$0.04{\pm}0.02$}

\newcommand{\dRDCVaRSimThree}{$0.32{\pm}0.02$}
\newcommand{\dRDCVaRSimFour}{$0.27{\pm}0.02$}
\newcommand{\dRDCVaRSimCplx}{$0.22{\pm}0.02$}
\newcommand{\dRDCVaRRnD}{$0.18{\pm}0.03$}
\newcommand{\dRDCVaRIPed}{$0.14{\pm}0.03$}

\newcommand{\dWoRDCVaRSimThree}{{\boldmath $0.48{\pm}0.02$}}
\newcommand{\dWoRDCVaRSimFour}{{\boldmath $0.43{\pm}0.02$}}
\newcommand{\dWoRDCVaRSimCplx}{{\boldmath $0.32{\pm}0.02$}}
\newcommand{\dWoRDCVaRRnD}{{\boldmath $0.34{\pm}0.02$}}
\newcommand{\dWoRDCVaRIPed}{{\boldmath $0.26{\pm}0.04$}}

\newcommand{\dNaiveCaWSimThree}{$19.8{\pm}2.0$}
\newcommand{\dNaiveCaWSimFour}{$28.4{\pm}2.7$}
\newcommand{\dNaiveCaWSimCplx}{$39.7{\pm}3.5$}
\newcommand{\dNaiveCaWRnD}{$44.6{\pm}5.8$}
\newcommand{\dNaiveCaWIPed}{$50.1{\pm}6.6$}

\newcommand{\dOWRCaWSimThree}{$11.4{\pm}1.6$}
\newcommand{\dOWRCaWSimFour}{$17.6{\pm}2.0$}
\newcommand{\dOWRCaWSimCplx}{$25.2{\pm}2.6$}
\newcommand{\dOWRCaWRnD}{$29.8{\pm}4.4$}
\newcommand{\dOWRCaWIPed}{$47.8{\pm}5.9$}

\newcommand{\dPFCaWSimThree}{$14.2{\pm}1.8$}
\newcommand{\dPFCaWSimFour}{$19.3{\pm}2.1$}
\newcommand{\dPFCaWSimCplx}{$23.1{\pm}2.3$}
\newcommand{\dPFCaWRnD}{$24.7{\pm}3.8$}
\newcommand{\dPFCaWIPed}{$28.6{\pm}4.5$}

\newcommand{\dGRUCaWSimThree}{$24.7{\pm}2.2$}
\newcommand{\dGRUCaWSimFour}{$36.5{\pm}2.9$}
\newcommand{\dGRUCaWSimCplx}{$50.8{\pm}3.7$}
\newcommand{\dGRUCaWRnD}{$48.4{\pm}4.6$}
\newcommand{\dGRUCaWIPed}{$51.2{\pm}5.8$}

\newcommand{\dRDCaWSimThree}{$7.0{\pm}1.3$}
\newcommand{\dRDCaWSimFour}{$14.6{\pm}1.8$}
\newcommand{\dRDCaWSimCplx}{$17.2{\pm}2.0$}
\newcommand{\dRDCaWRnD}{$22.4{\pm}3.4$}
\newcommand{\dRDCaWIPed}{$24.4{\pm}3.4$}

\newcommand{\dWoRDCaWSimThree}{{\boldmath $4.6{\pm}0.9$}}
\newcommand{\dWoRDCaWSimFour}{{\boldmath $8.5{\pm}1.3$}}
\newcommand{\dWoRDCaWSimCplx}{{\boldmath $11.4{\pm}1.6$}}
\newcommand{\dWoRDCaWRnD}{{\boldmath $14.1{\pm}2.9$}}
\newcommand{\dWoRDCaWIPed}{{\boldmath $17.0{\pm}3.2$}}

%% file: sections/experiments.tex
\section{Experiments}
\label{sec:experiments}

We evaluate \WoRD{} on five domains and five baselines. The domains split into two settings by whether goal-directed behavior is distinctive in the early prefix: \emph{non-complex}, with optimal trajectories that diverge early, and \emph{complex}, with prefixes that stay ambiguous for many timesteps. The central claim under test is that \WoRD{} preserves \multimodal{} beliefs through these ambiguous prefixes, which yields better performance when premature \unimodal{} commitment misrepresents the true posterior, while still converging onto $w^*$ once the trajectory becomes distinctive. The formulation also transfers across agent types without method-level changes, consistent with the data-source-agnostic training procedure of \cref{sec:training}.

\subsection{Setup}
\label{sec:exp-setup}

{\bf Domains.}
 We created two simulated domains by procedurally generating 2D worlds: \simWP{3} and \simWP{4} use three and four waypoints, respectively, while \simWPc{} (\simWPc*) uses a four-waypoint complex variant with longer ambiguous prefixes. Obstacle layouts and trajectories are randomized to produce ambiguous prefixes, the setting in which \multimodal{} belief preservation matters.
For real-world data, we use two human-driven datasets. From the rounD dataset~\citep{rounDdataset} we extracted the vehicle subset \rounDV{} (\rounDV*); from the inD dataset~\citep{bock2020ind} we extracted the pedestrian-agent subset \indPed{} (\indPed*).
For each real-world domain, we retrain the score network with the same two-stage procedure described in \cref{sec:training}. Locations covered, goal extraction, clustering hyperparameters, per-domain sampling rates, observation models, and episode counts are in \cref{app:datasets}. \simWP{3} and \simWP{4} are the non-complex domains; \simWPc{}, \rounDV{}, and \indPed{} are the complex domains with extended ambiguous prefixes.
\paragraph{Baselines.}
We compare \WoRD{} against five baselines spanning the main families of online goal-recognition methods, in the order of the failure-mode discussion in \cref{sec:intro}:
(1)~\textbf{Naive}, a distance-and-heading geometric heuristic. It serves as a floor: since it ignores dynamics, observation history, and goal structure, a recognizer that fails to outperform it is mismatched to the domain.
(2)~\textbf{Particle Filter (PF)}, structurally identical to \OWR{} but replacing the per-waypoint Gaussian filter with a particle-based posterior, using dynamics propagation and observation-weighted resampling. This isolates the contribution of the unimodal Gaussian assumption from the rest of the OWR pipeline.
(3)~\textbf{Online Waypoint Recognizer (\OWR{})}~\citep{guo-online-2024}, a UKF-per-waypoint filter whose precomputed reference trajectories are generated from a hand-specified dynamic model of the agent.
(4)~\textbf{GRU Classifier (GRU)}~\citep{cho_learning_2014}, trained with cross-entropy on the same trajectory data, the discriminative-learning alternative for goal recognition~\citep{chiari_goal_2023, serina_towards_2025} that sees the same data but commits to a single softmax per timestep with no generative model.
(5)~\textbf{Reverse Diffuser (RD)}~\citep{janner_planning_2022}, which uses the same \WoRD{} model but applies observation guidance \emph{post-hoc} rather than during denoising and re-initializes each chain from noise at every update; RD is thus a controlled instance of the predict-then-classify strategy.

{\bf Metrics.}
A goal recognizer faces two failure modes that argmax accuracy alone cannot separate. When a distinctive path is present, the optimal posterior peaks sharply on $w^*$ and the recognizer should converge to it. When the prefix is non-distinctive (optimal plans toward different goals overlap), the true posterior is \multimodal{}; \unimodal{} collapse is the failure of placing $\max_j p_t(w^j)$ on a single waypoint before the trajectory disambiguates the goal, locking the recognizer onto a commitment that may be wrong. We use four metrics chosen to expose both modes.
\textbf{Top-1} ($\uparrow$, fraction of timesteps with $\arg\max_j p_t(w^j) = w^*$) and \textbf{Brier} ($\downarrow$, $\tfrac{1}{T}\sum_{t,j}(p_t(w^j) - \mathbb{1}[w^j{=}w^*])^2$) test convergence and full-posterior calibration.
\textbf{CVaR\textsubscript{0.1}} ($\uparrow$, mean $p_t(w^*)$ on the worst $10\%$ of (episode, timestep) pairs) and \textbf{CaW} ($\downarrow$, fraction of pairs with $\max_j p_t(w^j) \geq 0.92$ on an incorrect waypoint) test the other failure mode: whether the recognizer keeps $w^*$ in its support through ambiguous prefixes and avoids \unimodal{} collapse, which argmax-style metrics underweight.
Extended results, including per-scene inD/rounD breakdown, are in \cref{app:extended_results}.

{\bf Protocol.}
For each method we run $5$ training seeds and evaluate each seed on $50$ held-out episodes per simulation domain and $100$ per real-world domain, reporting the mean $\pm\;1.96\times\mathrm{SE}$ across seeds (\Cref{tab:implementation-hyperparams}). Real-world episodes come from a leave-one-location-out cross-validation, so the test location is never seen during training. All baselines see the same observation history $\obshistory$ and candidate set $\waypointset$ at every timestep. Hardware, GRU retraining, ablation seeds, and per-method wall-clock are documented in \cref{app:implementation,app:extended_results,app:ablations}.
\subsection{Results}
\label{sec:exp-main}
Per-domain numbers are reported in \Cref{tab:sim-results} (simulation) and \Cref{tab:realworld-results} (real-world); the four-metric trade-off scatter plots (\cref{fig:results-pareto}) are in \cref{app:extended_results}.

\input{tables/sim_results}

\input{tables/realworld_results}

{\bf Overall.}
\WoRD{} is best on every CVaR$_{0.1}$ and CaW cell and on the Top-1 and Brier cells of all three complex domains; three of the four non-complex Top-1 and Brier cells are statistical ties with GRU under overlapping $95\%$ CIs, with \simWP{3} Top-1 the one cell where GRU leads outside the CI (\Cref{tab:sim-results,tab:realworld-results}).
The split between the argmax-style metrics and the multimodal-belief metrics follows from the mechanism: argmax wins require keeping $w^*$ in the support throughout the prefix, which only matters when prefixes become ambiguous, while CVaR$_{0.1}$ and CaW penalize \unimodal{} collapse independently of whether the eventual argmax is correct.

\paragraph{Non-complex setting.}
On \simWP{3} and \simWP{4}, goal-directed trajectories diverge early enough that an aggressive early commitment is usually correct: the clarifying signal arrives before a wrong hypothesis can compound into a persistent error.
Top-1 and Brier consequently reduce to ties among methods that converge by the end of the prefix. GRU leads \WoRD{} on Top-1 by $3.8$\,pp on \simWP{3} and $0.5$\,pp on \simWP{4} (\Cref{tab:sim-results}); only the \simWP{3} gap falls outside the $95\%$ CI. This is the one cell where a discriminative classifier cleanly beats \WoRD{}, because \simWP{3} prefixes disambiguate too early to reward belief preservation. Brier ties on both.
CaW exposes the cost: GRU's confident-and-wrong rate on \simWP{3} is \dGRUCaWSimThree{}, more than five times \WoRD{}'s \dWoRDCaWSimThree{}, because cross-entropy training commits confidently regardless of prefix ambiguity.
Here the argmax is correct often enough to mask this miscalibration.

\paragraph{Complex setting.}
\rounDV{} most clearly shows why \multimodal{} belief preservation matters.
Vehicles entering a roundabout follow an identical circular arc for most of the trajectory; the exit maneuver is the first and only distinctive part of the trajectory.
Throughout the shared arc the optimal posterior is \multimodal{}: a well-calibrated recognizer should distribute belief over all plausible exits and commit only as the exit signal emerges.
CVaR$_{0.1}$ directly tests whether a method maintains this distribution.
GRU's CVaR$_{0.1}$ on \rounDV{} is \dGRUCVaRRnD{}: the true exit receives almost no probability mass during the shared arc, because cross-entropy training has already driven the softmax to commit to one exit long before the trajectory disambiguates the goal.
\WoRD{}'s CVaR$_{0.1}$ is \dWoRDCVaRRnD{}, a more than six-fold improvement, because per-step score guidance keeps probability spread across plausible exits and updates proportionally as the trajectory disambiguates the goal.
CaW shows the downstream consequence: GRU emits a confident-and-wrong belief on \dGRUCaWRnD{} of (episode, timestep) pairs, while \WoRD{} exceeds the $0.92$ threshold on an incorrect hypothesis on only \dWoRDCaWRnD{}.

The same pattern holds across the full complex set.
\WoRD{} leads GRU on Top-1 by $+10.4$\,pp on \simWPc{}, $+7.9$\,pp on \rounDV{}, and $+8.6$\,pp on \indPed{}, gaps that are negligible in the non-complex domains and arise because correct final recognition requires keeping $w^*$ in the support throughout the ambiguous prefix.
On \indPed{}, where pedestrians share entry corridors before diverging to distinct destinations, GRU's CaW reaches \dGRUCaWIPed{}, on par with Naive's \dNaiveCaWIPed{} despite GRU's access to the full observation history; cross-entropy training has amplified overconfidence to the point where GRU's safety-critical error rate matches a method with no observation model at all.
\WoRD{} remains at \dWoRDCaWIPed{}, a $34$\,pp gap that Top-1 alone does not reveal.

\paragraph{Mechanism.}
The Reverse Diffuser (RD) shares architecture and training with \WoRD{} but applies the observation likelihood as post-hoc reweighting of unguided samples and re-initializes each denoising chain from noise at every update instead of warm-starting. The CaW gap between RD and \WoRD{} therefore measures the joint contribution of online guidance and warm-start.
That gap widens substantially in the complex setting: $2.4$\,pp on \simWP{3}, $6.1$\,pp on \simWP{4}, $5.8$\,pp on \simWPc{}, $8.3$\,pp on \rounDV{}, and $7.4$\,pp on \indPed{}.
CVaR$_{0.1}$ shows the same pattern: RD's worst-decile $p_t(w^*)$ on \rounDV{} is \dRDCVaRRnD{} compared to \WoRD{}'s \dWoRDCVaRRnD{}.
Both metrics tie the \multimodal{} belief advantage to online inference rather than to the architecture or prior, which the two methods share. The $\lambda$ rows of \Cref{tab:ablations} isolate guidance alone: at $\lambda = 0$, Top-1 falls to chance.

\paragraph{Transfer.}
\OWR{}'s Top-1 drops by $50$\,pp from \simWP{3} to \indPed{} (\dOWRTopOneSimThree{} to \dOWRTopOneIPed{}) and PF's by $20$\,pp (\dPFTopOneSimThree{} to \dPFTopOneIPed{}) as their hand-coded dynamics models meet real-world geometry, while \WoRD{}'s drop over the same span is $19$\,pp (\dWoRDTopOneSimThree{} to \dWoRDTopOneIPed{}), comparable in magnitude; the loss is attributable to distribution shift in the learned prior, whereas \OWR{} and PF lose to model misspecification.

\paragraph{Key finding.}
\WoRD{} is the only observer in our evaluation that improves on all four metrics jointly. It reduces the safety-critical confident-and-wrong rate by roughly $8$ percentage points relative to the strongest baseline (\Cref{tab:realworld-results}), without regressing on Top-1 accuracy. The advantage is largest on domains with naturally ambiguous prefixes (intersections, roundabouts), where filter-based observers commit early and discriminative classifiers grow overconfident. The Reverse Diffuser, which shares \WoRD{}'s score network but reweights unguided samples post-hoc, is the strongest baseline on CaW and trails \WoRD{} by $0.06$ Brier and $\sim$$8$ percentage points of confident-and-wrong rate. The safety-critical gain therefore comes from online inference rather than from the learned prior.

%% file: tables/sim_results.tex
\begin{table}[t]
    \centering
    \caption{\textbf{Simulation results.} Bold marks the best entry in each column. \simWP{3} and \simWP{4} are non-complex domains; \simWPc{} is the complex variant.}
    \label{tab:sim-results}
    \small
    \setlength{\tabcolsep}{5pt}
    \begin{tabular}{l ccc ccc}
        \toprule
        & \multicolumn{3}{c}{Top-1 (\%)\,$\uparrow$}
        & \multicolumn{3}{c}{Brier\,$\downarrow$\,$[0,2]$} \\
        \cmidrule(lr){2-4}\cmidrule(lr){5-7}
        Method & 3WP & 4WP & 4WPc & 3WP & 4WP & 4WPc \\
        \midrule
        Naive   & \dNaiveTopOneSimThree & \dNaiveTopOneSimFour & \dNaiveTopOneSimCplx & \dNaiveBrierSimThree & \dNaiveBrierSimFour & \dNaiveBrierSimCplx \\
        \OWR{}  & \dOWRTopOneSimThree   & \dOWRTopOneSimFour   & \dOWRTopOneSimCplx   & \dOWRBrierSimThree   & \dOWRBrierSimFour   & \dOWRBrierSimCplx \\
        PF      & \dPFTopOneSimThree    & \dPFTopOneSimFour    & \dPFTopOneSimCplx    & \dPFBrierSimThree    & \dPFBrierSimFour    & \dPFBrierSimCplx \\
        GRU     & \dGRUTopOneSimThree   & \dGRUTopOneSimFour   & \dGRUTopOneSimCplx   & \dGRUBrierSimThree   & \dGRUBrierSimFour   & \dGRUBrierSimCplx \\
        RD      & \dRDTopOneSimThree    & \dRDTopOneSimFour    & \dRDTopOneSimCplx    & \dRDBrierSimThree    & \dRDBrierSimFour    & \dRDBrierSimCplx \\
        \WoRD{} & \dWoRDTopOneSimThree  & \dWoRDTopOneSimFour  & \dWoRDTopOneSimCplx  & \dWoRDBrierSimThree  & \dWoRDBrierSimFour  & \dWoRDBrierSimCplx \\
        \midrule
        & \multicolumn{3}{c}{CVaR$_{0.1}$\,$\uparrow$}
        & \multicolumn{3}{c}{CaW (\%)\,$\downarrow$} \\
        \cmidrule(lr){2-4}\cmidrule(lr){5-7}
        Method & 3WP & 4WP & 4WPc & 3WP & 4WP & 4WPc \\
        \midrule
        Naive   & \dNaiveCVaRSimThree & \dNaiveCVaRSimFour & \dNaiveCVaRSimCplx & \dNaiveCaWSimThree & \dNaiveCaWSimFour & \dNaiveCaWSimCplx \\
        \OWR{}  & \dOWRCVaRSimThree   & \dOWRCVaRSimFour   & \dOWRCVaRSimCplx   & \dOWRCaWSimThree   & \dOWRCaWSimFour   & \dOWRCaWSimCplx \\
        PF      & \dPFCVaRSimThree    & \dPFCVaRSimFour    & \dPFCVaRSimCplx    & \dPFCaWSimThree    & \dPFCaWSimFour    & \dPFCaWSimCplx \\
        GRU     & \dGRUCVaRSimThree   & \dGRUCVaRSimFour   & \dGRUCVaRSimCplx   & \dGRUCaWSimThree   & \dGRUCaWSimFour   & \dGRUCaWSimCplx \\
        RD      & \dRDCVaRSimThree    & \dRDCVaRSimFour    & \dRDCVaRSimCplx    & \dRDCaWSimThree    & \dRDCaWSimFour    & \dRDCaWSimCplx \\
        \WoRD{} & \dWoRDCVaRSimThree  & \dWoRDCVaRSimFour  & \dWoRDCVaRSimCplx  & \dWoRDCaWSimThree  & \dWoRDCaWSimFour  & \dWoRDCaWSimCplx \\
        \bottomrule
    \end{tabular}
\end{table}

%% file: tables/realworld_results.tex
\begin{table}[t]
    \centering
    \caption{\textbf{Real-world results.} Bold marks the best entry in each column. \rounDV{} aggregates the three roundabout locations; \indPed{} aggregates the four intersections. Per-scene splits are in \cref{app:per-scene}.}
    \label{tab:realworld-results}
    \footnotesize
    \setlength{\tabcolsep}{4pt}
    \resizebox{\textwidth}{!}{%
    \begin{tabular}{l cc cc cc cc}
        \toprule
        & \multicolumn{2}{c}{Top-1 (\%)\,$\uparrow$}
        & \multicolumn{2}{c}{Brier\,$\downarrow$\,$[0,2]$}
        & \multicolumn{2}{c}{CVaR$_{0.1}$\,$\uparrow$}
        & \multicolumn{2}{c}{CaW (\%)\,$\downarrow$} \\
        \cmidrule(lr){2-3}\cmidrule(lr){4-5}\cmidrule(lr){6-7}\cmidrule(lr){8-9}
        Method
        & rounD-V & inD-Ped
        & rounD-V & inD-Ped
        & rounD-V & inD-Ped
        & rounD-V & inD-Ped \\
        \midrule
        Naive
        & \dNaiveTopOneRnD & \dNaiveTopOneIPed
        & \dNaiveBrierRnD  & \dNaiveBrierIPed
        & \dNaiveCVaRRnD   & \dNaiveCVaRIPed
        & \dNaiveCaWRnD    & \dNaiveCaWIPed \\
        \OWR{}
        & \dOWRTopOneRnD & \dOWRTopOneIPed
        & \dOWRBrierRnD  & \dOWRBrierIPed
        & \dOWRCVaRRnD   & \dOWRCVaRIPed
        & \dOWRCaWRnD    & \dOWRCaWIPed \\
        PF
        & \dPFTopOneRnD & \dPFTopOneIPed
        & \dPFBrierRnD  & \dPFBrierIPed
        & \dPFCVaRRnD   & \dPFCVaRIPed
        & \dPFCaWRnD    & \dPFCaWIPed \\
        GRU
        & \dGRUTopOneRnD & \dGRUTopOneIPed
        & \dGRUBrierRnD  & \dGRUBrierIPed
        & \dGRUCVaRRnD   & \dGRUCVaRIPed
        & \dGRUCaWRnD    & \dGRUCaWIPed \\
        RD
        & \dRDTopOneRnD & \dRDTopOneIPed
        & \dRDBrierRnD  & \dRDBrierIPed
        & \dRDCVaRRnD   & \dRDCVaRIPed
        & \dRDCaWRnD    & \dRDCaWIPed \\
        \WoRD{}
        & \dWoRDTopOneRnD & \dWoRDTopOneIPed
        & \dWoRDBrierRnD  & \dWoRDBrierIPed
        & \dWoRDCVaRRnD   & \dWoRDCVaRIPed
        & \dWoRDCaWRnD    & \dWoRDCaWIPed \\
        \bottomrule
    \end{tabular}%
    }
\end{table}

%% file: sections/discussion.tex
\section{Conclusion}
\label{sec:discussion}
We present \WoRD{}, which casts online goal recognition as posterior inference under a learned diffusion prior: each candidate waypoint anchors an independent denoising chain, observation guidance enters analytically through the gradient of the measurement model, and warm-starting keeps per-update cost compatible with online operation. Across three simulated and two real-world domains, \WoRD{} is the only observer that improves on all four metrics jointly. It preserves calibrated \multimodal{} beliefs through ambiguous prefixes where filter- and classifier-based observers collapse to a single mode or grow overconfident, and it lowers the confident-and-wrong rate by roughly $8$ percentage points relative to the strongest baseline on real-world data.

{\bf Limitations.}
Three assumptions limit \WoRD{}'s applicability. The candidate waypoint set $\waypointset$ must be specified in advance: the method recognizes goals from a known finite set and cannot discover new ones at inference time. This closed-set assumption is standard in goal recognition and shared with \OWR{}; when it is violated, \WoRD{} degrades toward uncertainty, whereas \OWR{} commits confidently to a wrong goal (\cref{app:candidate-sensitivity}). All real-world results hold out the test location, so deployment does not require training data from the target site. Symbolic goal recognition over discrete action domains is complementary to our continuous-trajectory setting and outside our scope. The measurement model $h$ must be known and differentiable; sensor models with non-differentiable gates (rejected detections, hard occlusion masks) require a smoothed surrogate. Finally, although warm-starting reduces per-update cost by a factor of $\numsteps / \diffstep^*$, \WoRD{} remains more expensive per observation than a Kalman filter step (\cref{app:runtime}); on hardware-constrained platforms the appropriate operating point trades off the cost of confident-and-wrong predictions against inference latency.

{\bf Future Directions.}
\WoRD{} currently operates on planar position trajectories; extending to full 3D state with attitude and angular velocity is a question of training data and network capacity rather than method-level changes. Open-set goal discovery, in which $\waypointset$ is augmented from observations, would relax the most restrictive limitation above; the conditional score network already supports continuous goal coordinates, so the remaining bottleneck is detecting new candidates. Latent diffusion would extend \WoRD{} to image- or video-based intent inference without changing the inference principle, but raises two open problems: guidance is informative only if progress toward a goal varies smoothly in the learned latent geometry, and calibration must survive the encoder. Finally, the \multimodal{} posteriors that \WoRD{} maintains are the input that risk-sensitive planners need to reason about branching futures and deceptive agents. 

{\bf Broader Impacts}.
Better-calibrated goal recognition supports safer autonomous planning: planners can defer commitment under ambiguity rather than collapse to a single hypothesis and replan when the agent diverges. The same framework applied to passive trajectory streams could enable behavioral surveillance, and a planner that anticipates pedestrian intent could in principle plan adversarially against humans. Responsible deployment therefore requires consent frameworks for the data sources that feed the score network, and oversight of the planners that consume \WoRD{}'s posterior.

%% file: sections/appendix.tex
\crefalias{section}{appendix}
\crefname{appendix}{Appendix}{Appendices}
\Crefname{appendix}{Appendix}{Appendices}

\input{sections/appendix/bayes}
\input{sections/appendix/algorithm}
\input{sections/appendix/implementation}
\input{sections/appendix/datasets}
\input{sections/appendix/ablations}

\input{sections/appendix/extended_results}

%% file: sections/appendix/bayes.tex
\section{Bayes' Rule on Observations}
\label{app:bayes}

Bayes' rule applied to $p(\traj_{\diffstep} \mid w^j, \obshistory)$ gives
\begin{equation}
\label{eq:bayes-joint-A}
p(\traj_{\diffstep} \mid w^j, \obshistory) \;=\; \frac{p(\obshistory \mid \traj_{\diffstep}, w^j)\, p(\traj_{\diffstep} \mid w^j)}{p(\obshistory \mid w^j)}.
\end{equation}
Taking the logarithm and then the gradient with respect to $\traj_{\diffstep}$ removes the evidence term $\log p(\obshistory \mid w^j)$, which is constant in $\traj_{\diffstep}$:
\begin{equation}
\label{eq:bayes-grad-A}
\nabla_{\traj_{\diffstep}} \log p(\traj_{\diffstep} \mid w^j, \obshistory) \;=\; \nabla_{\traj_{\diffstep}} \log p(\traj_{\diffstep} \mid w^j) \;+\; \nabla_{\traj_{\diffstep}} \log p(\obshistory \mid \traj_{\diffstep}, w^j).
\end{equation}
The measurement model of \cref{eq:obs-model} depends on the trajectory alone, so $p(\obshistory \mid \traj_{\diffstep}, w^j) = p(\obshistory \mid \traj_{\diffstep})$, recovering \cref{eq:score-decomp} with $\lambda = 1$. In practice, following classifier guidance~\citep{dhariwal_diffusion_2021}, we treat $\lambda$ as a tunable guidance scale: $\lambda > 1$ sharpens the posterior toward observation-consistent trajectories at the cost of prior coverage, while $\lambda < 1$ relaxes it. The Bayesian derivation no longer holds exactly under $\lambda \ne 1$; we report sensitivity in \cref{app:ablations}.

%% file: sections/appendix/algorithm.tex
\section{Online inference algorithm}
\label{app:algorithm}
\Cref{alg:word} shows the full \WoRD{} online process.
\input{figures/word_algorithm_inference}
The pseudocode maps line-by-line to the equations of \cref{sec:method}. Lines~1 and 2 implement the warm-start branch: at the first observation ($k{=}1$) every candidate's chain is initialized from pure noise and runs through the full $\numsteps$ reverse steps, while at every subsequent step the cached estimate $\hat{\traj}_{k-1}^j$ is re-noised to $\diffstep^*$ via the closed-form forward marginal of \cref{eq:warmstart}. Lines~3 through 10 form the guided reverse chain, repeated independently per candidate $w^j$: line~5 evaluates the stage-2 noise predictor $\noisepredict$ from \cref{sec:training}; line~6 computes the observation-likelihood gradient $\mathbf{g}^j$ of \cref{eq:obs-likelihood} by backpropagation through the measurement model $h$, exploiting the differentiability assumption of \cref{sec:problem}; line~7 forms the standard DDPM mean of \cref{eq:reverse}; and line~8 draws the next iterate from a Gaussian whose mean is shifted by the guidance term $\sigma_{\diffstep}^2 \mathbf{g}^j$, implementing the score modification of \cref{eq:score-decomp}. Line~11 reweights the candidates via Bayes' rule with a uniform candidate prior and the single-sample marginal-likelihood estimator of \cref{eq:mc-likelihood} (the $S{=}1$ case), evaluating the observation likelihood at the denoised endpoint $\traj_0^j$ and normalizing across $j$ to produce a proper distribution. Line~12 caches this endpoint per candidate so that the next observation can re-noise from a near-data-manifold trajectory rather than from pure noise. The output is the full waypoint posterior $\{p(w^j \mid \obshistory)\}_{j=1}^N$ over $\waypointset$, which the multimodal-belief metrics of \cref{sec:experiments} (CVaR$_{0.1}$, CaW, Brier) consume directly rather than reduced to an argmax.

%% file: figures/word_algorithm_inference.tex
\begin{algorithm}[H]
\caption{\WoRD{} Online Inference with Warm-Start}
\label{alg:word}
\begin{algorithmic}[1]
\REQUIRE Observations $\obshistory$, previous estimate $\hat{\traj}_{k-1}$ (or $\emptyset$ if $k=1$), noise predictor $\noisepredict$, guidance scale $\lambda$, warm-start step $\diffstep^*$
\STATE \textbf{if} $k=1$ \textbf{then} $\traj_{\numsteps}^j \sim \mathcal{N}(\mathbf{0},\mathbf{I})$ for all $j$, $\;\diffstep_{\mathrm{s}} \leftarrow \numsteps$
\STATE \textbf{else} $\traj_{\diffstep^*}^j \sim q(\hat{\traj}_{k-1},\diffstep^*)$ for all $j$, $\;\diffstep_{\mathrm{s}} \leftarrow \diffstep^*$ \COMMENT{Eq.~\ref{eq:warmstart}}
\FOR{$\diffstep = \diffstep_{\mathrm{s}}$ \textbf{downto} $1$}
    \FOR{each $w^j \in \waypointset$}
        \STATE $\boldsymbol{\epsilon} \leftarrow \noisepredict(\traj_{\diffstep}^j, w^j, \diffstep, \obshistory)$ \COMMENT{noise prediction}
        \STATE $\mathbf{g}^j \leftarrow \nabla_{\traj_{\diffstep}^j} \log p(\obshistory \mid \traj_{\diffstep}^j)$ \COMMENT{Eq.~\ref{eq:obs-likelihood}, via autograd}
        \STATE $\boldsymbol{\mu}^j \leftarrow \text{DDPMStep}(\traj_{\diffstep}^j, \boldsymbol{\epsilon})$ \COMMENT{Eq.~\ref{eq:reverse}}
        \STATE $\traj_{\diffstep-1}^j \sim \mathcal{N}(\boldsymbol{\mu}^j + \sigma_{\diffstep}^2\,\mathbf{g}^j,\;\sigma_{\diffstep}^2\mathbf{I})$
    \ENDFOR
\ENDFOR
\STATE $p(w^j \mid \obshistory) \;\propto\; p(w^j) \cdot \exp\!\left(-\textstyle\sum_{t=1}^k\left\|\obs_t - h(\traj_0^j(t))\right\|_{\measnoise^{-1}}^2\right)$ for all $j$ \COMMENT{Eq.~\ref{eq:mc-likelihood}, normalize over $j$}
\STATE $\hat{\traj}_{k}^j \leftarrow \traj_0^j$ for all $j$ \COMMENT{store per-goal; used as warm-start at step $k{+}1$}
\ENSURE Posterior $\{p(w^j \mid \obshistory)\}_{j=1}^N$ over the candidate set $\waypointset$
\end{algorithmic}
\end{algorithm}

%% file: sections/appendix/implementation.tex
\section{Implementation details}
\label{app:implementation}

\paragraph{Architecture.}
The score network is a Temporal U-Net~\citep{janner_planning_2022} operating on trajectory tensors of shape $(T, d_{\text{in}})$ with $T{=}100$ and $d_{\text{in}}{=}6$ input channels per timestep: the diffused position $(\traj_x, \traj_y)$, the observed position $(\obs_x, \obs_y)$, the observation mask, and the heading $\theta$. The encoder uses three downsampling scales with channel widths $[32, 64, 128]$ and two residual blocks per scale; the decoder mirrors the encoder with skip connections. Each residual block is a 1D convolution (kernel size~3) followed by GroupNorm (8 groups) and a SiLU nonlinearity, with FiLM conditioning~\citep{perez_film_2018} injected by a learned linear projection from the concatenation of the waypoint embedding $w^j \in \mathbb{R}^2$ and a sinusoidal diffusion-step embedding of dimension~$64$. The output head projects back to $(T, 2)$, predicting noise on the $(x, y)$ channels only. The full network has $2.2$M parameters.

\paragraph{Hyperparameters.}
\Cref{tab:implementation-hyperparams} lists the hyperparameters used throughout the paper for both training and online inference. Inference values are the canonical defaults; ablations on $\diffstep^*$, $\lambda$, $S$, and the score-network architecture are reported in \cref{app:ablations}.

\begin{table}[h]
\centering
\caption{Implementation hyperparameters used in \WoRD{} experiments.}
\label{tab:implementation-hyperparams}
\small
\begin{tabular}{lll}
\toprule
\textbf{Component} & \textbf{Hyperparameter} & \textbf{Value} \\
\midrule
\multirow{4}{*}{Diffusion}
  & Trajectory horizon $T$ & $100$ \\
  & Number of steps $\numsteps$ & $100$ \\
  & $\beta$ schedule & linear, $\beta \in [10^{-4},\, 2{\times}10^{-2}]$ \\
\midrule
\multirow{6}{*}{Network}
  & Architecture & Temporal U-Net \\
  & Channels per scale & $[32, 64, 128]$ \\
  & Residual blocks per scale & $2$ \\
  & Time embedding dim & $64$ \\
  & Conditioning & FiLM on $w^j$ \\
  & Total parameters & $2.2$M \\
\midrule
\multirow{4}{*}{Training (both stages)}
  & Optimizer & Adam, lr $=10^{-4}$ \\
  & Batch size & $64$ \\
  & Gradient clip ($\ell_2$) & $1.0$ \\
  & Epochs (stage 1 / stage 2) & $200$ / $100$ \\
\midrule
\multirow{2}{*}{Stage 2 augmentation}
  & Prefix length sampling & $k = \lfloor \rho T \rfloor,\; \rho \sim \mathcal{U}(0.3,\, 0.9)$ \\
  & Observation noise std & $0.1$ \\
\midrule
\multirow{4}{*}{Inference}
  & Warm-start step $\diffstep^*$ & $20$ \\
  & Guidance scale $\lambda$ & $1.0$ \\
  & Samples per candidate $S$ & $50$ \\
  & Posterior temperature $\tau$ & $1$ \\
\midrule
\multirow{3}{*}{Evaluation}
  & Training seeds & $5$ \\
  & Episodes per seed (sim.) & $50$ \\
  & Episodes per seed (real-world) & $100$ \\
\bottomrule
\end{tabular}
\end{table}

\paragraph{Compute.}
All training and evaluation use a single NVIDIA L40 GPU. Two-stage training completes in under one day on the simulation dataset and in a few hours per recording fold on inD and rounD. Online inference cost is dominated by the U-Net forward pass and scales linearly in the warm-start step $\diffstep^*$ and the sample count $S$.

\subsection{Runtime}
\label{app:runtime}
\Cref{tab:runtime} reports the per-update wall-clock of every method, averaged over $10{,}000$ updates, with all methods of a scenario timed on the same GPU. The cost of the classical filters grows with the number of candidates, whereas both diffusion-based methods are dominated by a fixed per-update denoising cost that is insensitive to it.

\begin{table}[h]
\centering
\caption{Per-update wall-clock in milliseconds, mean over $10{,}000$ updates. All methods within a scenario are timed on the same GPU.}
\label{tab:runtime}
\small
\begin{tabular}{lccc}
\toprule
\textbf{Method} & \textbf{2WP} & \textbf{3WP} & \textbf{4WP} \\
\midrule
Naive            & $0.036$ & $0.036$ & $0.038$ \\
\OWR{} (UKF)     & $0.96$  & $1.40$  & $1.89$  \\
Particle Filter  & $0.85$  & $1.26$  & $1.69$  \\
GRU              & $2.70$  & $2.72$  & $2.72$  \\
\rowcolor{gray!15}
\WoRD{} (ours)   & $6.8$   & $10.3$  & $12.2$  \\
Reverse Diffuser & $111$   & $169$   & $199$   \\
\bottomrule
\end{tabular}
\end{table}

Iterative denoising is more expensive per update than a Kalman or particle-filter step; \WoRD{}'s per-update cost of $6.8$ to $12.2$\,ms is nonetheless of the same order as the $10$\,ms observation budget and therefore compatible with online operation. Goal recognition also tolerates a slower update rate than low-level control, since the destination of a pedestrian or vehicle remains stable over seconds and the belief is consumed by a planning or decision layer. The implementation is not optimized for inference speed: it runs in full precision with an uncompiled sampling loop and without step distillation, so \cref{tab:runtime} is an upper bound on the cost of \WoRD{}.

\WoRD{} is roughly $16\times$ faster than the Reverse Diffuser, which re-initializes every denoising chain from noise at each update; warm-starting from the previous estimate removes this cost. In the 2WP ablation setting (\cref{tab:ablations}), a cold start ($\diffstep^*=100$) is about $5\times$ slower than the default $\diffstep^*=20$ and also less accurate. The cost can be reduced further at a modest loss in accuracy: relative to the default configuration, $S=25$ samples per candidate cost about $0.6\times$ at $91.2$ Top-1 (against $93.1$), and the MLP score network costs about $0.7\times$ at $89.2$ Top-1.

%% file: sections/appendix/datasets.tex
\section{Datasets}
\label{app:datasets}

\subsection{Real-World Datasets}
\label{app:datasets-real}

\input{figures/datasets_overview.tex}

\paragraph{Goal extraction.}
For both real-world domains (inD pedestrians and rounD vehicles), candidate goals are extracted from the terminal positions of agent trajectories using per-agent-group DBSCAN clustering, with $\varepsilon{=}3.5$\,m and $\textsc{min\_samples}{=}5$. \Cref{fig:datasets-overview} shows one location from each domain with the recovered candidate set overlaid on the recording's aerial. Per-domain sampling rates, observation models, and episode counts are summarized below.

\subsection{Simulation}
\label{app:datasets-sim}

\input{figures/datasets_simulation.tex}

\paragraph{Synthetic scenarios.}
The synthetic environments used for diverse-dataset training and the candidate-count scaling experiments are configurable $10{\times}10$\,m worlds with a differential-drive agent and static circular obstacles. In each panel of \cref{fig:datasets-simulation}, stars denote the candidate goals $\{w^{j}\}$, the white circle marks the shared start at $(5,\,1.5)$, and gray discs are the obstacles. The 3WP scenario~(\cref{fig:datasets-sim-3wp}) places three top-row goals around a single central obstacle. The simple 4WP scenario~(\cref{fig:datasets-sim-4wp}) retains that obstacle but compresses the goal set into a single horizontal row, isolating the effect of candidate count. The complex 4WP variant~(\cref{fig:datasets-sim-4wp-complex}) replaces the central obstacle with four scattered discs that induce two sequential lateral forks before the agent reaches the goal row, isolating the effect of obstacle layout. The full diverse training pool covers $50$ environments sampled from this configuration space.

%% file: figures/datasets_overview.tex
\begin{figure*}[t]
\centering
\definecolor{wcolzero}{HTML}{1F78B4}
\definecolor{wcolone} {HTML}{33A02C}
\definecolor{wcoltwo} {HTML}{E31A1C}
\definecolor{wcolthree}{HTML}{FF7F00}
\tikzset{
  wpstar/.style 2 args={
    star, star points=5, star point ratio=2.4,
    minimum size=11pt, line width=0.7pt,
    draw=#1, fill=#2, inner sep=0pt,
  },
  wplabel/.style={
    fill=black!8, rounded corners=2pt,
    inner sep=2pt, font=\scriptsize\bfseries,
  },
}

\begin{subfigure}[t]{0.52\linewidth}
\centering
\begin{tikzpicture}
  \node[anchor=south west, inner sep=0] (img) at (0,0) {%
    \includegraphics[width=\linewidth]
                    {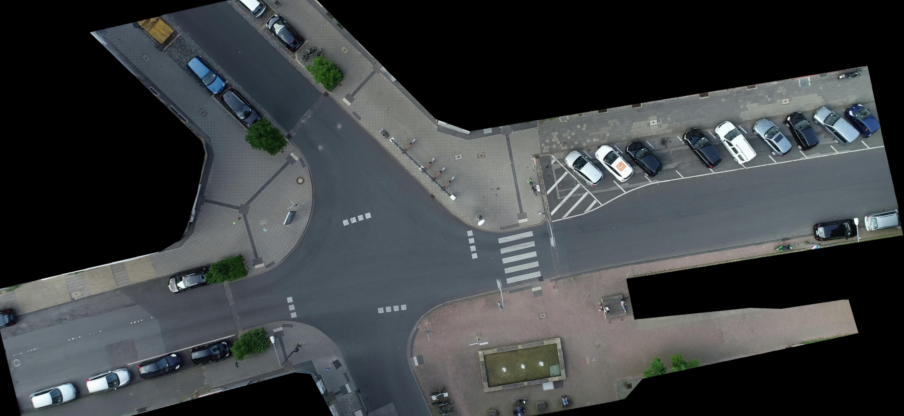}%
  };
  \begin{scope}[x={(img.south east)}, y={(img.north west)}]
    \node[wpstar={wcolzero}{wcolzero!18}] (s0) at (0.1273, 0.9123) {};
    \node[wplabel, anchor=north west, text=wcolzero]
      at ($(s0)+(0.018,-0.030)$) {$w^{0}$};
    \node[wpstar={wcolone}{wcolone!18}] (s1) at (0.9185, 0.7893) {};
    \node[wplabel, anchor=north east, text=wcolone]
      at ($(s1)+(-0.018,-0.030)$) {$w^{1}$};
    \node[wpstar={wcoltwo}{wcoltwo!18}] (s2) at (0.6377, 0.1210) {};
    \node[wplabel, anchor=south west, text=wcoltwo]
      at ($(s2)+(0.018,0.030)$) {$w^{2}$};
    \node[wpstar={wcolthree}{wcolthree!18}] (s3) at (0.0202, 0.2769) {};
    \node[wplabel, anchor=south west, text=wcolthree]
      at ($(s3)+(0.018,0.030)$) {$w^{3}$};
  \end{scope}
\end{tikzpicture}
\caption{inD location~2 (Bendplatz, pedestrians). $W{=}4$ candidate goals.}
\label{fig:datasets-ind-loc2}
\end{subfigure}\hfill
\begin{subfigure}[t]{0.46\linewidth}
\centering
\begin{tikzpicture}
  \node[anchor=south west, inner sep=0] (img) at (0,0) {%
    \includegraphics[width=\linewidth]
                    {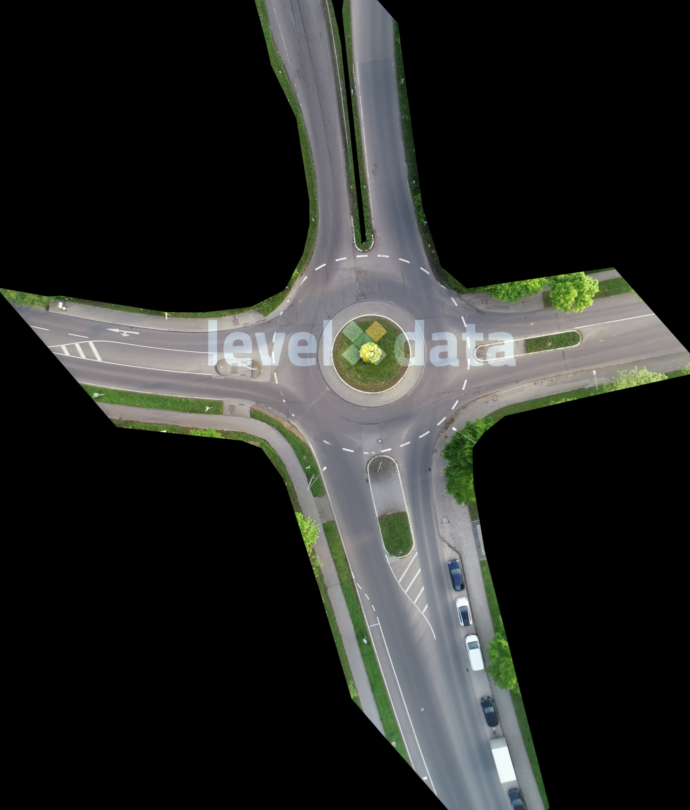}%
  };
  \begin{scope}[x={(img.south east)}, y={(img.north west)}]
    \node[wpstar={wcolzero}{wcolzero!18}] (s0) at (0.5493, 0.9376) {};
    \node[wplabel, anchor=north, text=wcolzero]
      at ($(s0)+(0.000,-0.025)$) {$w^{0}$};
    \node[wpstar={wcolone}{wcolone!18}] (s1) at (0.9242, 0.5727) {};
    \node[wplabel, anchor=east, text=wcolone]
      at ($(s1)+(-0.025,0.000)$) {$w^{1}$};
    \node[wpstar={wcoltwo}{wcoltwo!18}] (s2) at (0.5971, 0.1879) {};
    \node[wplabel, anchor=south, text=wcoltwo]
      at ($(s2)+(0.000,0.025)$) {$w^{2}$};
    \node[wpstar={wcolthree}{wcolthree!18}] (s3) at (0.0738, 0.5977) {};
    \node[wplabel, anchor=west, text=wcolthree]
      at ($(s3)+(0.025,0.000)$) {$w^{3}$};
  \end{scope}
\end{tikzpicture}
\caption{rounD location~1 (Thiergarten, motor vehicles). $W{=}4$ candidate goals.}
\label{fig:datasets-round-loc1}
\end{subfigure}

\caption{\textbf{Representative locations from the two real-world domains, with
DBSCAN-extracted candidate goals overlaid.}
Stars mark cluster centroids of agent terminal positions. Aerial backgrounds are recordings rec28 (inD) and
rec00 (rounD).}
\label{fig:datasets-overview}
\end{figure*}
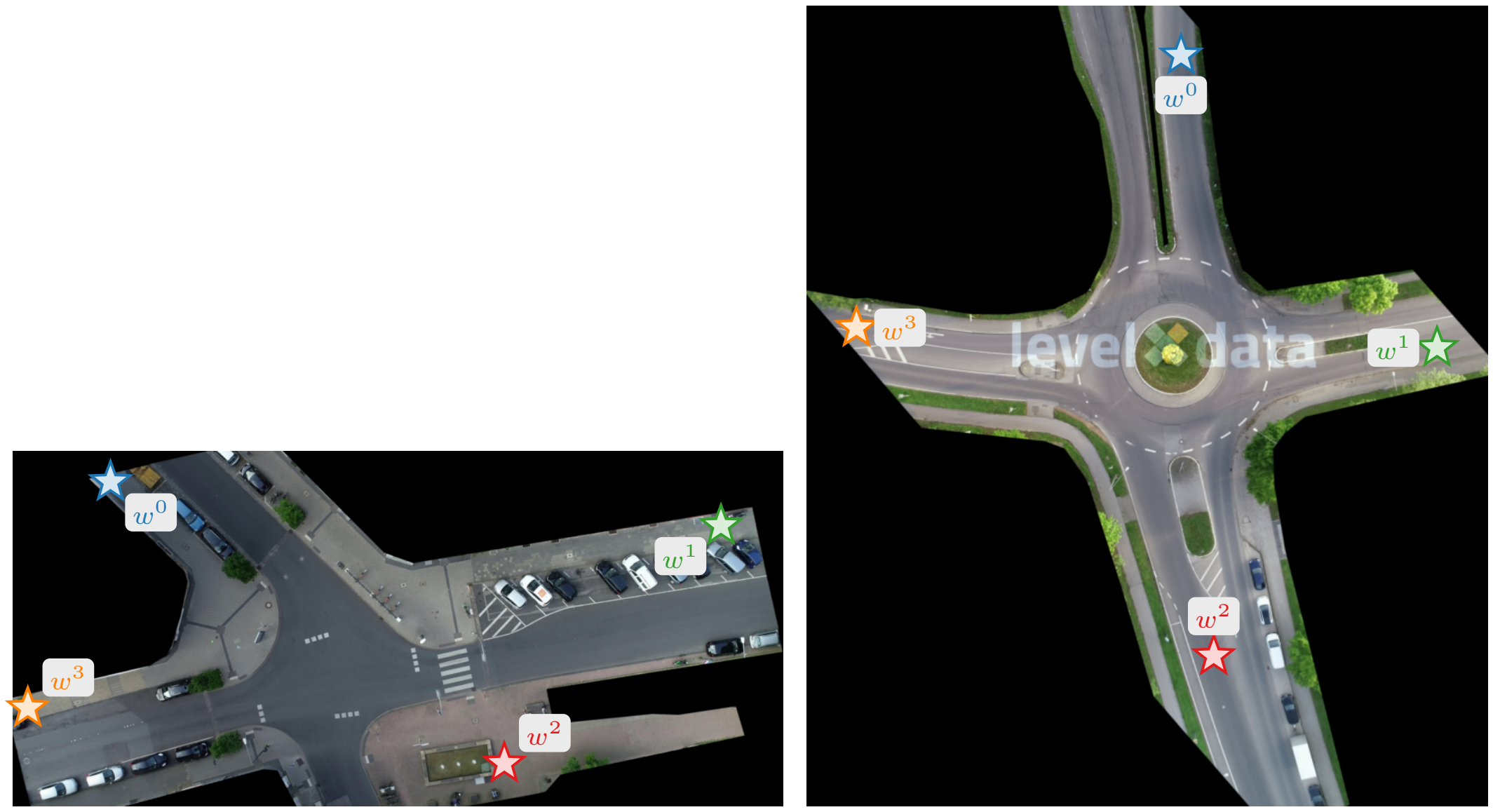

%% file: figures/datasets_simulation.tex
\begin{figure*}[t]
\centering
\definecolor{wcolzero} {HTML}{1F78B4}
\definecolor{wcolone}  {HTML}{33A02C}
\definecolor{wcoltwo}  {HTML}{E31A1C}
\definecolor{wcolthree}{HTML}{FF7F00}

\def\BoardW{10}
\def\BoardH{10}
\def\StartX{5}
\def\StartY{1.5}

\tikzset{
  simframe/.style={x=0.40cm, y=0.40cm},
  simgrid/.style={gray!12, very thin},
  simborder/.style={black!55, line width=0.5pt},
  simobs/.style={fill=gray!28, draw=black!55, line width=0.4pt},
  simstart/.style={fill=white, draw=black, line width=0.7pt},
  wpstar/.style 2 args={
    star, star points=5, star point ratio=2.3,
    minimum size=9pt, line width=0.6pt,
    draw=#1, fill=#2, inner sep=0pt,
  },
  wplabel/.style={
    fill=black!8, rounded corners=1.5pt,
    inner sep=1.5pt, font=\tiny\bfseries,
  },
}

\newcommand{\drawWorld}{%
  \draw[simgrid] (0,0) grid[step=2] (\BoardW,\BoardH);
  \draw[simborder] (0,0) rectangle (\BoardW,\BoardH);
}
\newcommand{\drawStart}{%
  \draw[simstart] (\StartX,\StartY) circle (0.20);
  \node[font=\tiny, anchor=north, text=black!75]
    at (\StartX,\StartY-0.30) {start};
}
\newcommand{\drawObs}[3]{\fill[simobs] (#1,#2) circle (#3);}
\newcommand{\drawWP}[7]{%
  \node[wpstar={#2}{#2!18}] (s#1) at (#3,#4) {};
  \node[wplabel, anchor=#5, text=#2]
    at ($(s#1)+(#6,#7)$) {$w^{#1}$};
}

\begin{subfigure}[t]{0.32\linewidth}
\centering
\begin{tikzpicture}[simframe]
  \begin{scope}\clip (-0.05,-0.05) rectangle (\BoardW+0.05,\BoardH+0.05);
    \drawWorld
    \drawObs{5.0}{5.0}{1.0}
    \drawWP{0}{wcolzero}{3.0}{8.0}{south east}{-0.10}{0.30}
    \drawWP{1}{wcolone} {7.0}{8.0}{south west}{ 0.10}{0.30}
    \drawWP{2}{wcoltwo} {5.0}{9.0}{south}     { 0.00}{0.35}
    \drawStart
  \end{scope}
\end{tikzpicture}
\caption{3 goals (simple). One central obstacle.}
\label{fig:datasets-sim-3wp}
\end{subfigure}\hfill
\begin{subfigure}[t]{0.32\linewidth}
\centering
\begin{tikzpicture}[simframe]
  \begin{scope}\clip (-0.05,-0.05) rectangle (\BoardW+0.05,\BoardH+0.05);
    \drawWorld
    \drawObs{5.0}{5.0}{1.0}
    \drawWP{0}{wcolzero} {2.00}{8.0}{south east}{-0.10}{0.30}
    \drawWP{1}{wcolone}  {4.33}{8.0}{south east}{-0.10}{0.30}
    \drawWP{2}{wcoltwo}  {5.67}{8.0}{south west}{ 0.10}{0.30}
    \drawWP{3}{wcolthree}{8.00}{8.0}{south west}{ 0.10}{0.30}
    \drawStart
  \end{scope}
\end{tikzpicture}
\caption{4 goals (simple). Closely-spaced top-row candidates discriminated by~$x$.}
\label{fig:datasets-sim-4wp}
\end{subfigure}\hfill
\begin{subfigure}[t]{0.32\linewidth}
\centering
\begin{tikzpicture}[simframe]
  \begin{scope}\clip (-0.05,-0.05) rectangle (\BoardW+0.05,\BoardH+0.05);
    \drawWorld
    \drawObs{5.0}{3.0}{0.5}
    \drawObs{3.0}{5.0}{0.5}
    \drawObs{7.0}{5.0}{0.5}
    \drawObs{5.0}{6.5}{0.5}
    \drawWP{0}{wcolzero} {2.00}{8.0}{south east}{-0.10}{0.30}
    \drawWP{1}{wcolone}  {4.33}{8.0}{south east}{-0.10}{0.30}
    \drawWP{2}{wcoltwo}  {5.67}{8.0}{south west}{ 0.10}{0.30}
    \drawWP{3}{wcolthree}{8.00}{8.0}{south west}{ 0.10}{0.30}
    \drawStart
  \end{scope}
\end{tikzpicture}
\caption{4 goals (complex). Two sequential center forks plus lateral flanks force compounded commitment decisions.}
\label{fig:datasets-sim-4wp-complex}
\end{subfigure}

\caption{\textbf{Synthetic simulation scenarios for diverse-dataset training and
candidate-count scaling.}}
\label{fig:datasets-simulation}
\end{figure*}
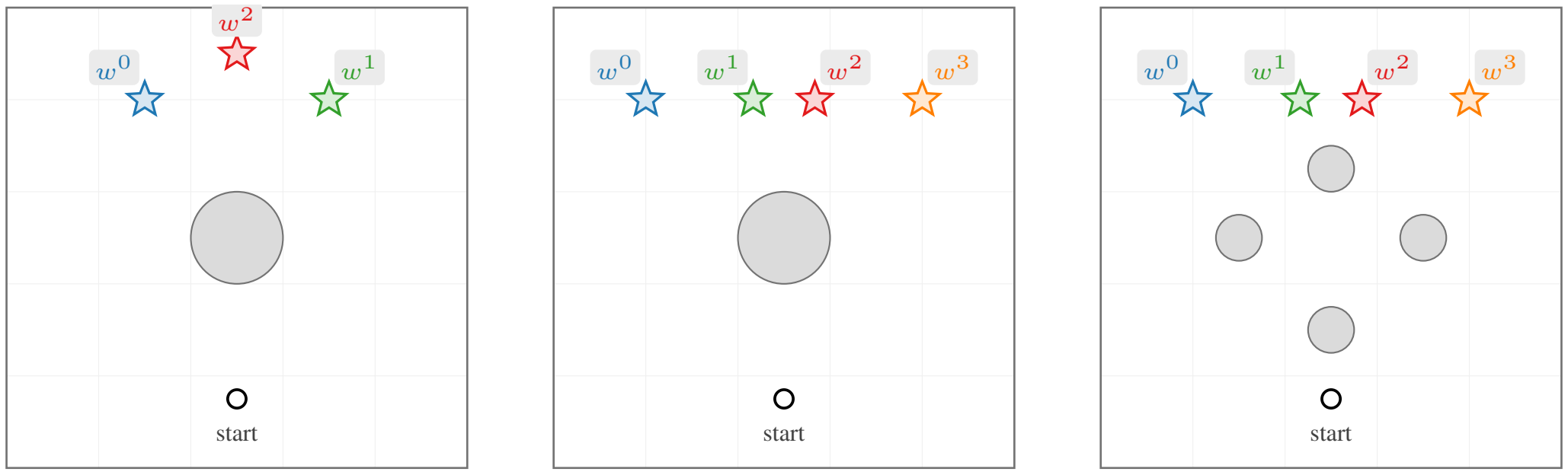

%% file: sections/appendix/ablations.tex
\section{Ablation studies}
\label{app:ablations}

These ablations probe sensitivity to design choices; the primary evidence for our main claims is the five-domain evaluation in \cref{sec:experiments}. All runs use the 2WP toy environment: a $10\times10$\,m world with two goals at $(3, 8)$ and $(7, 8)$, a single circular obstacle of radius $1$ at $(5, 5)$, and a fixed start at $(5, 1.5)$. The symmetric two-goal layout admits a closed-form analytical posterior, which we use as a ground-truth reference to cross-check implementations before scaling to the full evaluation set. Each cell reports the mean and bootstrap $95\%$ CI over (episode, timestep) pairs from $n=5$ training seeds. The direction of each effect below transfers to the five primary domains, though the absolute magnitudes do not.

\begin{table}[h]
    \caption{Ablation studies on \WoRD{} design choices, all runs on the 2WP environment. Default configuration shaded. CIs are bootstrap $95\%$ over $(\text{episode},\,t)$ pairs from $n=5$ seeds.}
    \label{tab:ablations}
    \centering
    \small
    \begin{tabular}{llcc}
        \toprule
        \textbf{Factor} & \textbf{Value} & \textbf{Top-1} $\uparrow$ & \textbf{Brier} $\downarrow$ \\
        \midrule
        \multirow{4}{*}{Warm-start $\diffstep^*$}
            & $0$ (no re-noise)                  & $76.2{\pm}2.4$                    & $0.20{\pm}0.02$ \\
            & $10$                               & $88.7{\pm}1.8$                    & $0.10{\pm}0.01$ \\
            & \cellcolor{gray!15}$20$ (default)  & \cellcolor{gray!15}$93.1{\pm}1.4$ & \cellcolor{gray!15}$0.07{\pm}0.01$ \\
            & $100$ (cold start)                 & $89.4{\pm}1.7$                    & $0.09{\pm}0.01$ \\
        \midrule
        \multirow{4}{*}{Guidance $\lambda$}
            & $0.0$ (prior only)                 & $49.1{\pm}2.1$                    & $0.32{\pm}0.02$ \\
            & $0.5$                              & $81.6{\pm}2.0$                    & $0.13{\pm}0.02$ \\
            & \cellcolor{gray!15}$1.0$ (default) & \cellcolor{gray!15}$93.4{\pm}1.4$ & \cellcolor{gray!15}$0.07{\pm}0.01$ \\
            & $5.0$                              & $58.7{\pm}2.3$                    & $0.13{\pm}0.02$ \\
        \midrule
        \multirow{2}{*}{Architecture}
            & \cellcolor{gray!15}Temporal U-Net  & \cellcolor{gray!15}$93.1{\pm}1.4$ & \cellcolor{gray!15}$0.07{\pm}0.01$ \\
            & MLP                                & $89.2{\pm}1.9$                    & $0.09{\pm}0.02$ \\
        \midrule
        \multirow{4}{*}{Samples $S$}
            & $10$                               & $86.4{\pm}2.0$                    & $0.12{\pm}0.02$ \\
            & $25$                               & $91.2{\pm}1.6$                    & $0.09{\pm}0.01$ \\
            & \cellcolor{gray!15}$50$ (default)  & \cellcolor{gray!15}$93.1{\pm}1.4$ & \cellcolor{gray!15}$0.07{\pm}0.01$ \\
            & $100$                              & $93.6{\pm}1.5$                    & $0.06{\pm}0.01$ \\
        \bottomrule
    \end{tabular}
\end{table}

\subsection{Sensitivity to the candidate set}
\label{app:candidate-sensitivity}
The closed-set assumption of \cref{sec:problem} can fail at deployment in two ways: $\waypointset$ may contain irrelevant candidates, or it may omit the agent's true destination. We test both cases on 2WP with the trained 2WP model and without retraining, which is possible because the network is conditioned on waypoint coordinates rather than on a fixed set of class indices (\cref{sec:prior}). The GRU classifier cannot be evaluated in either case, since its output layer is fixed to the training goals.

\paragraph{Irrelevant candidates.}
We add $2$ or $6$ distractor candidates placed away from the paths to the two true goals. As the candidate set grows, the accuracy of \WoRD{} falls and its peak confidence falls with it (\cref{tab:distractors}): six distractors cost $18$ points of accuracy and $31$ points of peak confidence. This coupling is the property a downstream planner relies on, since a belief that loses accuracy while retaining its confidence would mislead it. The model-based baselines retain a confidence of $1.0$ in this setting, because their hand-specified reference trajectories rule out off-path candidates entirely; the same mechanism makes them confidently wrong when the true goal is absent.

\begin{table}[h]
\centering
\caption{Irrelevant distractor candidates added to the 2WP candidate set. Accuracy, CaW, and mean peak confidence of \WoRD{}.}
\label{tab:distractors}
\small
\begin{tabular}{lccc}
\toprule
\textbf{Candidate set} & \textbf{Acc} $\uparrow$ & \textbf{CaW} $\downarrow$ & \textbf{Max-conf} \\
\midrule
\rowcolor{gray!15}
$2$ goals (control)       & $0.93$ & $0.00$ & $0.91$ \\
$+2$ distractors ($4$)    & $0.82$ & $0.24$ & $0.63$ \\
$+6$ distractors ($8$)    & $0.75$ & $0.30$ & $0.60$ \\
\bottomrule
\end{tabular}
\end{table}

\paragraph{True goal removed.}
For each test episode we remove the true goal, so that every observer receives the remaining goal and two off-path candidates and no candidate explains the observed motion. Since every candidate is wrong, any confidence is misplaced. \Cref{tab:missing-goal} reports the confidence each observer places on its top candidate at the final step, over $200$ episodes ($100$ per removed goal).

\begin{table}[h]
\centering
\caption{True goal removed from the candidate set (2WP, $200$ episodes). With no correct candidate, lower confidence and higher entropy are better. Values at the final step, averaged over the two removed-goal conditions. Entropy uses the natural logarithm, so its maximum for three candidates is $\ln 3 \approx 1.10$ nats.}
\label{tab:missing-goal}
\small
\begin{tabular}{lccc}
\toprule
\textbf{Method} & \textbf{Mean max-conf} $\downarrow$ & \textbf{Episodes with max-conf} $\geq 0.9$ $\downarrow$ & \textbf{Entropy (nats)} $\uparrow$ \\
\midrule
\rowcolor{gray!15}
\WoRD{} (ours)  & $0.80$ & $3\%$   & $0.58$ \\
Particle Filter & $0.82$ & $38\%$  & $0.45$ \\
Naive           & $0.98$ & $95\%$  & $0.04$ \\
\OWR{}          & $1.00$ & $100\%$ & $0.00$ \\
\bottomrule
\end{tabular}
\end{table}

\WoRD{} is the least confident observer when the true goal is absent (\cref{tab:missing-goal}). \OWR{} renormalizes its near-deterministic likelihoods over the remaining candidates and commits to a wrong goal with full confidence in every episode, whereas \WoRD{} weights each candidate by how well the learned prior explains the observed motion, so that no candidate receives confident mass when none fits. The same weighting explains the distractor result: implausible candidates are down-weighted without receiving zero mass, so adding them dilutes the confidence of \WoRD{}.

%% file: sections/appendix/extended_results.tex
\section{Extended experimental results}
\label{app:extended_results}

\subsection{Pareto trade-offs across metrics}
\label{app:pareto}
\Cref{fig:results-pareto} reads the per-cell rankings of \Cref{tab:sim-results,tab:realworld-results} as four metric trade-offs, with one point per (method, domain) on each panel. Color identifies the method; marker shape identifies the domain (legend below the figure). The four panels separate the two failure modes the metrics target: panels (a) and (b) ask whether a method converges and is calibrated when it does; panels (c) and (d) ask whether it preserves multimodal belief through ambiguous prefixes. Methods that win one pair but not the other, most visibly GRU on (b) but not (c), are the reason we report four metrics rather than Top-1 alone.

\import{figures/}{results_pareto.tex}

\textbf{Panel (a), Top-1 vs.\ CaW (accuracy vs.\ safety):} \WoRD{} reaches the desirable corner on every domain, and the orange cluster sits well below the GRU red cluster on the three complex domains (triangle, diamond, pentagon).
\textbf{Panel (b), Top-1 vs.\ Brier (accuracy vs.\ calibration):} on the non-complex sim domains (circle, square) GRU and \WoRD{} are statistical ties; on the three complex domains \WoRD{} dominates by a Brier margin that grows with prefix ambiguity.
\textbf{Panel (c), CVaR$_{0.1}$ vs.\ CaW (tail success vs.\ tail failure):} the most direct test of \multimodal{} belief preservation. \WoRD{} sits uniformly to the right of every other method, and the gap is largest on rounD-V (diamond), where the shared circular arc keeps the optimal posterior multimodal for most of the trajectory.
\textbf{Panel (d), Brier vs.\ CaW with reversed x-axis:} the bottom-right corner combines low calibration error with low \unimodal{} collapse; the GRU cluster (red) sits high on CaW despite competitive Brier, the signature of cross-entropy overconfidence.
Per-cell numerical values are in \Cref{tab:sim-results,tab:realworld-results}.

\subsection{Per-scene inD/rounD breakdown}
\label{app:per-scene}
\Cref{tab:results-per-scene} decomposes the headline real-world Top-1 numbers by recording location: inD pedestrians at four signalized-intersection locations (Loc~1--4 in standard inD nomenclature, left half) and rounD vehicle traffic at three roundabout locations (Loc~0--2, right half). The rightmost ``Avg'' column reproduces the headline Top-1 from \Cref{tab:realworld-results}; per-location standard deviations across the $5$ training seeds range from $\pm 2.4$ to $\pm 4.2$ for \WoRD{} and from $\pm 3.1$ to $\pm 5.3$ for the baselines. The method ordering, \WoRD{} $>$ RD $>$ GRU $>$ PF $>$ \OWR{} $>$ Naive, is preserved at every location, so the aggregate ranking is not driven by a single easy or hard scene. Rank-preserving variation across locations spans $6$--$10$ Top-1 points within each dataset and reflects geometry density: rounD Loc~0 (the densest roundabout) and inD Loc~3--4 (the largest multi-arm intersections) are the hardest scenes for every method.

\import{figures/}{results_per_scene.tex}

%% file: figures/results_pareto.tex
\definecolor{cNaive}{HTML}{B0B0B0}
\definecolor{cOWR}{HTML}{4C72B0}
\definecolor{cPF}{HTML}{55A868}
\definecolor{cGRU}{HTML}{C44E52}
\definecolor{cRD}{HTML}{8172B2}
\definecolor{cWoRD}{HTML}{DD8452}

\newcommand{\sccpt}[3]{\addplot[only marks, color=#1, mark=#2, mark size=2.4pt, forget plot] coordinates {(#3)};}
\newcommand{\sccwd}[2]{\addplot[only marks, color=cWoRD, mark=#1, mark size=2.8pt, line width=0.8pt, forget plot] coordinates {(#2)};}

\begin{figure*}[t]
    \centering
    \begin{tikzpicture}
        \begin{groupplot}[
            group style={group size=2 by 2, horizontal sep=1.4cm, vertical sep=1.3cm},
            width=0.43\linewidth, height=4.6cm,
            grid=both, grid style={gray!15, line width=0.3pt},
            label style={font=\scriptsize}, tick label style={font=\scriptsize},
            title style={font=\scriptsize, yshift=-2pt},
        ]

        \nextgroupplot[
            xlabel={Top-1 (\%) $\rightarrow$}, ylabel={CaW (\%) $\leftarrow$},
            xmin=24, xmax=92, ymin=0, ymax=56,
            title={\textbf{(a) Accuracy vs.\ safety}},
        ]
        \sccpt{cNaive}{*}{64.2,19.8}        \sccpt{cNaive}{square*}{55.6,28.4}
        \sccpt{cNaive}{triangle*}{38.4,39.7}\sccpt{cNaive}{diamond*}{32.8,44.6}
        \sccpt{cNaive}{pentagon*}{27.4,50.1}
        \sccpt{cOWR}{*}{81.4,11.4}          \sccpt{cOWR}{square*}{75.2,17.6}
        \sccpt{cOWR}{triangle*}{58.7,25.2}  \sccpt{cOWR}{diamond*}{53.1,29.8}
        \sccpt{cOWR}{pentagon*}{31.4,47.8}
        \sccpt{cPF}{*}{74.8,14.2}           \sccpt{cPF}{square*}{76.8,19.3}
        \sccpt{cPF}{triangle*}{64.2,23.1}   \sccpt{cPF}{diamond*}{62.7,24.7}
        \sccpt{cPF}{pentagon*}{54.2,28.6}
        \sccpt{cGRU}{*}{88.4,24.7}          \sccpt{cGRU}{square*}{82.9,36.5}
        \sccpt{cGRU}{triangle*}{67.8,50.8}  \sccpt{cGRU}{diamond*}{63.4,48.4}
        \sccpt{cGRU}{pentagon*}{57.1,51.2}
        \sccpt{cRD}{*}{84.3,7.0}            \sccpt{cRD}{square*}{80.7,14.6}
        \sccpt{cRD}{triangle*}{70.4,17.2}   \sccpt{cRD}{diamond*}{64.9,22.4}
        \sccpt{cRD}{pentagon*}{58.6,24.4}
        \sccwd{*}{84.6,4.6}                 \sccwd{square*}{82.4,8.5}
        \sccwd{triangle*}{78.2,11.4}        \sccwd{diamond*}{71.3,14.1}
        \sccwd{pentagon*}{65.7,17.0}

        \nextgroupplot[
            xlabel={Top-1 (\%) $\rightarrow$}, ylabel={Brier $\leftarrow$},
            xmin=24, xmax=92, ymin=0.10, ymax=0.70,
            title={\textbf{(b) Accuracy vs.\ calibration}},
        ]
        \sccpt{cNaive}{*}{64.2,0.41}        \sccpt{cNaive}{square*}{55.6,0.49}
        \sccpt{cNaive}{triangle*}{38.4,0.59}\sccpt{cNaive}{diamond*}{32.8,0.63}
        \sccpt{cNaive}{pentagon*}{27.4,0.67}
        \sccpt{cOWR}{*}{81.4,0.17}          \sccpt{cOWR}{square*}{75.2,0.24}
        \sccpt{cOWR}{triangle*}{58.7,0.34}  \sccpt{cOWR}{diamond*}{53.1,0.42}
        \sccpt{cOWR}{pentagon*}{31.4,0.48}
        \sccpt{cPF}{*}{74.8,0.19}           \sccpt{cPF}{square*}{76.8,0.23}
        \sccpt{cPF}{triangle*}{64.2,0.30}   \sccpt{cPF}{diamond*}{62.7,0.31}
        \sccpt{cPF}{pentagon*}{54.2,0.36}
        \sccpt{cGRU}{*}{88.4,0.12}          \sccpt{cGRU}{square*}{82.9,0.16}
        \sccpt{cGRU}{triangle*}{67.8,0.27}  \sccpt{cGRU}{diamond*}{63.4,0.32}
        \sccpt{cGRU}{pentagon*}{57.1,0.39}
        \sccpt{cRD}{*}{84.3,0.14}           \sccpt{cRD}{square*}{80.7,0.18}
        \sccpt{cRD}{triangle*}{70.4,0.23}   \sccpt{cRD}{diamond*}{64.9,0.27}
        \sccpt{cRD}{pentagon*}{58.6,0.31}
        \sccwd{*}{84.6,0.13}                \sccwd{square*}{82.4,0.17}
        \sccwd{triangle*}{78.2,0.19}        \sccwd{diamond*}{71.3,0.22}
        \sccwd{pentagon*}{65.7,0.25}

        \nextgroupplot[
            xlabel={CVaR$_{0.1}$ $\rightarrow$}, ylabel={CaW (\%) $\leftarrow$},
            xmin=0.0, xmax=0.55, ymin=0, ymax=56,
            title={\textbf{(c) Tail success vs.\ tail failure}},
        ]
        \sccpt{cNaive}{*}{0.06,19.8}        \sccpt{cNaive}{square*}{0.04,28.4}
        \sccpt{cNaive}{triangle*}{0.03,39.7}\sccpt{cNaive}{diamond*}{0.02,44.6}
        \sccpt{cNaive}{pentagon*}{0.02,50.1}
        \sccpt{cOWR}{*}{0.18,11.4}          \sccpt{cOWR}{square*}{0.13,17.6}
        \sccpt{cOWR}{triangle*}{0.08,25.2}  \sccpt{cOWR}{diamond*}{0.06,29.8}
        \sccpt{cOWR}{pentagon*}{0.05,47.8}
        \sccpt{cPF}{*}{0.26,14.2}           \sccpt{cPF}{square*}{0.22,19.3}
        \sccpt{cPF}{triangle*}{0.17,23.1}   \sccpt{cPF}{diamond*}{0.14,24.7}
        \sccpt{cPF}{pentagon*}{0.11,28.6}
        \sccpt{cGRU}{*}{0.13,24.7}          \sccpt{cGRU}{square*}{0.10,36.5}
        \sccpt{cGRU}{triangle*}{0.07,50.8}  \sccpt{cGRU}{diamond*}{0.05,48.4}
        \sccpt{cGRU}{pentagon*}{0.04,51.2}
        \sccpt{cRD}{*}{0.32,7.0}            \sccpt{cRD}{square*}{0.27,14.6}
        \sccpt{cRD}{triangle*}{0.22,17.2}   \sccpt{cRD}{diamond*}{0.18,22.4}
        \sccpt{cRD}{pentagon*}{0.14,24.4}
        \sccwd{*}{0.48,4.6}                 \sccwd{square*}{0.43,8.5}
        \sccwd{triangle*}{0.32,11.4}        \sccwd{diamond*}{0.34,14.1}
        \sccwd{pentagon*}{0.26,17.0}

        \nextgroupplot[
            xlabel={Brier $\leftarrow$}, ylabel={CaW (\%) $\leftarrow$},
            xmin=0.10, xmax=0.70, ymin=0, ymax=56,
            x dir=reverse,
            title={\textbf{(d) Calibration vs.\ safety}},
            legend to name=paretolegend,
            legend style={
                draw=none, font=\footnotesize, column sep=0.55em, row sep=2pt,
                /tikz/every even column/.append style={column sep=0.4em},
            },
            legend columns=6,
        ]
        \sccpt{cNaive}{*}{0.41,19.8}        \sccpt{cNaive}{square*}{0.49,28.4}
        \sccpt{cNaive}{triangle*}{0.59,39.7}\sccpt{cNaive}{diamond*}{0.63,44.6}
        \sccpt{cNaive}{pentagon*}{0.67,50.1}
        \sccpt{cOWR}{*}{0.17,11.4}          \sccpt{cOWR}{square*}{0.24,17.6}
        \sccpt{cOWR}{triangle*}{0.34,25.2}  \sccpt{cOWR}{diamond*}{0.42,29.8}
        \sccpt{cOWR}{pentagon*}{0.48,47.8}
        \sccpt{cPF}{*}{0.19,14.2}           \sccpt{cPF}{square*}{0.23,19.3}
        \sccpt{cPF}{triangle*}{0.30,23.1}   \sccpt{cPF}{diamond*}{0.31,24.7}
        \sccpt{cPF}{pentagon*}{0.36,28.6}
        \sccpt{cGRU}{*}{0.12,24.7}          \sccpt{cGRU}{square*}{0.16,36.5}
        \sccpt{cGRU}{triangle*}{0.27,50.8}  \sccpt{cGRU}{diamond*}{0.32,48.4}
        \sccpt{cGRU}{pentagon*}{0.39,51.2}
        \sccpt{cRD}{*}{0.14,7.0}            \sccpt{cRD}{square*}{0.18,14.6}
        \sccpt{cRD}{triangle*}{0.23,17.2}   \sccpt{cRD}{diamond*}{0.27,22.4}
        \sccpt{cRD}{pentagon*}{0.31,24.4}
        \sccwd{*}{0.13,4.6}                 \sccwd{square*}{0.17,8.5}
        \sccwd{triangle*}{0.19,11.4}        \sccwd{diamond*}{0.22,14.1}
        \sccwd{pentagon*}{0.25,17.0}

        \addlegendimage{area legend, fill=cNaive, draw=cNaive!60!black} \addlegendentry{Naive}
        \addlegendimage{area legend, fill=cOWR,   draw=cOWR!60!black}   \addlegendentry{\OWR{}}
        \addlegendimage{area legend, fill=cPF,    draw=cPF!60!black}    \addlegendentry{PF}
        \addlegendimage{area legend, fill=cGRU,   draw=cGRU!60!black}   \addlegendentry{GRU}
        \addlegendimage{area legend, fill=cRD,    draw=cRD!60!black}    \addlegendentry{RD}
        \addlegendimage{area legend, fill=cWoRD,  draw=cWoRD!60!black}  \addlegendentry{\WoRD{}}
        \addlegendimage{only marks, mark=*,         mark size=2.4pt, color=gray!55!black} \addlegendentry{Sim 3WP}
        \addlegendimage{only marks, mark=square*,   mark size=2.4pt, color=gray!55!black} \addlegendentry{Sim 4WP}
        \addlegendimage{only marks, mark=triangle*, mark size=2.4pt, color=gray!55!black} \addlegendentry{Sim 4WPc}
        \addlegendimage{only marks, mark=diamond*,  mark size=2.4pt, color=gray!55!black} \addlegendentry{rounD-V}
        \addlegendimage{only marks, mark=pentagon*, mark size=2.4pt, color=gray!55!black} \addlegendentry{inD-Ped}

        \end{groupplot}
        \node[anchor=north] at ($(group c1r2.south east)!0.5!(group c2r2.south west) + (0,-0.85cm)$)
            {\pgfplotslegendfromname{paretolegend}};
    \end{tikzpicture}
    \caption{\textbf{Four metric trade-offs across the five evaluation domains.} Each panel scatters one point per (method, domain) for the six methods over the five domains. Colour identifies the method; marker shape identifies the domain (legend below). \WoRD{} points carry a thicker outline. The desirable corner is bottom-right in every panel; panel (d) reverses the x-axis so lower Brier still points right.}
    \label{fig:results-pareto}
\end{figure*}
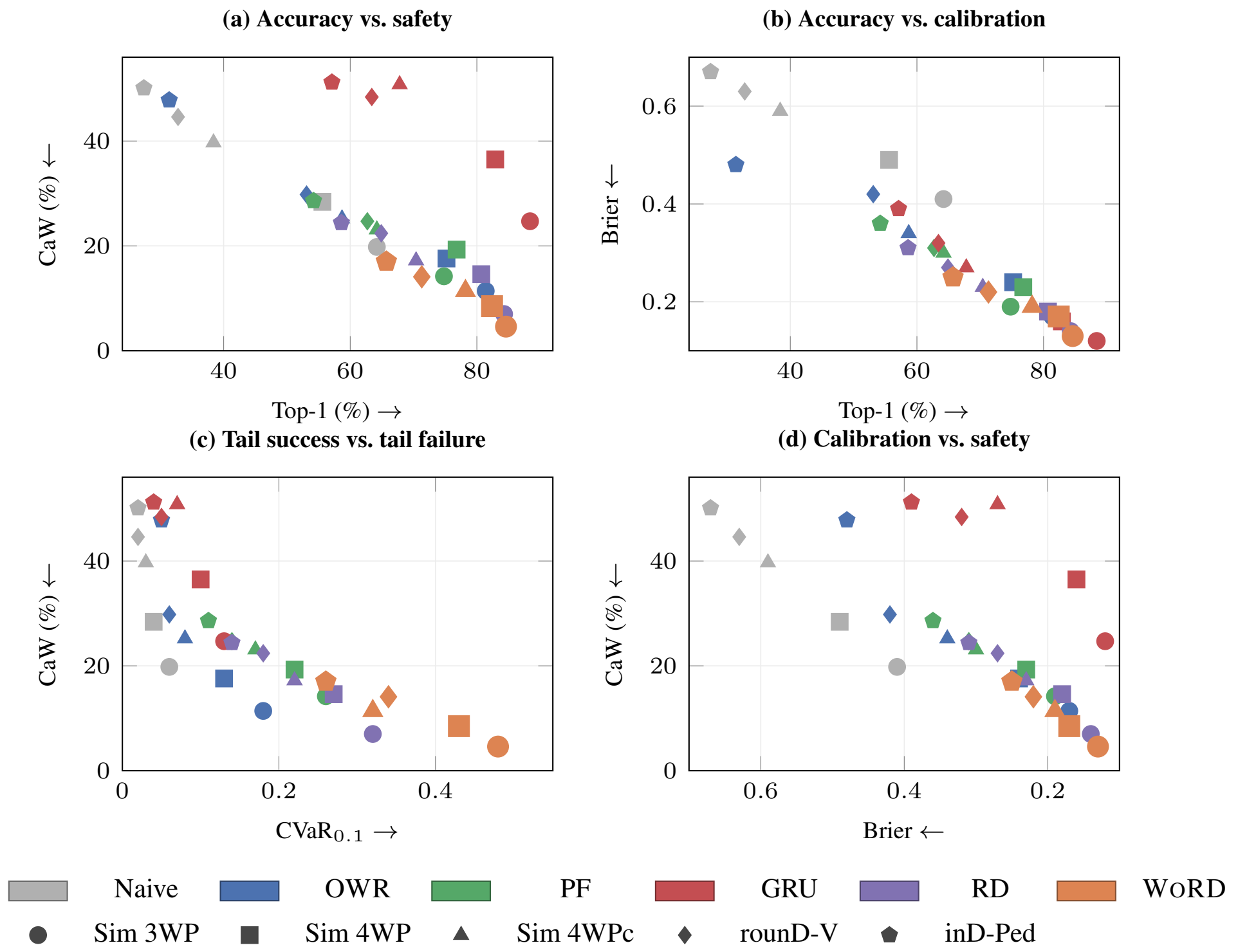

%% file: figures/results_per_scene.tex
\begin{table}[t]
    \centering
    \caption{Top-1 accuracy decomposed by recording location for the two real-world domains. \textbf{Left:} inD pedestrians, four signalized-intersection locations. \textbf{Right:} rounD vehicle traffic, three roundabout locations.}
    \label{tab:results-per-scene}
    \footnotesize
    \setlength{\tabcolsep}{4pt}
    \begin{minipage}[t]{0.55\linewidth}
        \centering
        \textbf{inD pedestrians: Top-1 (\%)\,$\uparrow$}\\[2pt]
        \begin{tabular}{l ccccc}
            \toprule
            Method  & Loc 1 & Loc 2 & Loc 3 & Loc 4 & Avg \\
            \midrule
            Naive   & 31.2 & 28.5 & 25.4 & 24.5 & 27.4 \\
            \OWR{}  & 35.8 & 32.9 & 28.8 & 28.1 & 31.4 \\
            PF      & 58.1 & 55.7 & 51.8 & 51.2 & 54.2 \\
            GRU     & 61.4 & 58.8 & 54.5 & 53.7 & 57.1 \\
            RD      & 63.2 & 60.5 & 56.0 & 54.7 & 58.6 \\
            \WoRD{} & \textbf{70.4} & \textbf{67.5} & \textbf{63.0} & \textbf{61.8} & \textbf{65.7} \\
            \bottomrule
        \end{tabular}
    \end{minipage}\hfill
    \begin{minipage}[t]{0.42\linewidth}
        \centering
        \textbf{rounD: Top-1 (\%)\,$\uparrow$}\\[2pt]
        \begin{tabular}{l cccc}
            \toprule
            Method  & Loc 0 & Loc 1 & Loc 2 & Avg \\
            \midrule
            Naive   & 29.0 & 35.6 & 33.8 & 32.8 \\
            \OWR{}  & 49.0 & 56.7 & 53.6 & 53.1 \\
            PF      & 58.8 & 66.4 & 62.9 & 62.7 \\
            GRU     & 59.5 & 67.1 & 63.6 & 63.4 \\
            RD      & 60.9 & 68.7 & 65.1 & 64.9 \\
            \WoRD{} & \textbf{67.4} & \textbf{74.8} & \textbf{71.7} & \textbf{71.3} \\
            \bottomrule
        \end{tabular}
    \end{minipage}
\end{table}